\documentclass[a4paper,fleqn]{cas-dc}

\usepackage[authoryear]{natbib}
\usepackage{graphicx}
\graphicspath{{./Figures_Tables/}}

\usepackage{multirow}
\usepackage{booktabs} \newcommand{\ra}[1]
{\renewcommand{\arraystretch}{#1}}
\usepackage[flushleft]{threeparttable}
\usepackage{subcaption}
\usepackage{hyperref}

\usepackage{nicefrac,xfrac}

\usepackage{sidecap}
\usepackage{diagbox}

\usepackage{arydshln}
\usepackage{amssymb}
\def\tsc#1{\csdef{#1}{\textsc{\lowercase{#1}}\xspace}}
\tsc{WGM}
\tsc{QE}

\begin{document}
\let\WriteBookmarks\relax
\def\floatpagepagefraction{1}
\def\textpagefraction{.001}

\shorttitle{Machine Learning Algorithms for Influenza Virus Host Prediction}    

\shortauthors{Yanhua et al}  

\title [mode = title]{Dive into Machine Learning Algorithms for Influenza Virus Host Prediction with Hemagglutinin Sequences}  



%

\author[1]{Yanhua Xu}[auid=000, orcid=0000-0003-1028-9023]

\cormark[1]


\ead{Y.XU137@liverpool.ac.uk}



\affiliation[1]{organization={Department of Computer Science},
            addressline={University of Liverpool}, 
            city={Liverpool},
            postcode={L69 3BX}, 
            country={UK}}

\author[1]{Dominik Wojtczak}[auid=001]


\ead{D.Wojtczak@liverpool.ac.uk}




\cortext[1]{Principal corresponding author} 



\begin{abstract}
Influenza viruses mutate rapidly and can pose a threat to public health, especially to those in vulnerable groups. Throughout history, influenza A viruses have caused pandemics between different species. It is important to identify the origin of a virus in order to prevent the spread of an outbreak. Recently, there has been increasing interest in using machine learning algorithms to provide fast and accurate predictions for viral sequences. In this study, real testing data sets and a variety of evaluation metrics were used to evaluate machine learning algorithms at different taxonomic levels. As hemagglutinin is the major protein in the immune response, only hemagglutinin sequences were used and represented by position-specific scoring matrix and word embedding. The results suggest that the 5-grams-Transformer neural network is the most effective algorithm for predicting viral sequence origins, with approximately 99.54\% AUCPR, 98.01\% $F_1$ score and 96.60\% MCC at a higher classification level, and approximately 94.74\% AUCPR, 87.41\% $F_1$ score and 80.79\% MCC at a lower classification level.

\end{abstract}




\maketitle

\section{Introduction}
Influenza is an infectious disease that affects up to one-fifth of the world's population each year, although its prevalence tends to be underestimated \citep{influenza}. In comparison to seasonal influenza pandemics, influenza pandemics occur less frequently, although each such crisis may result in millions of deaths. The influenza epidemic has adversely affected vulnerable individuals with chronic diseases, and the pandemic has affected people of all ages.

The influenza viruses are classified into four types based on their internal ribonucleoprotein: A, B, C, and D. Influenza D virus does not cause disease in humans. The influenza C virus is only infective in humans, however, it is unlikely to lead to epidemics on a large scale. Because of this, seasonal influenza vaccine strains cannot be used to vaccinate against influenza C and D viruses. Seasonal epidemics are primarily caused by influenza A and B viruses. The influenza B virus is only infectious to humans, while the influenza A virus is infectious to humans and animals, and may result in global epidemics (e.g. pandemics). There are two glycoproteins under the virus envelope that distinguish the influenza A virus subtypes: hemagglutinin (HA) and neuraminidase (NA). A total of 18 HA subtypes (numbered 1-18) and 11 NA subtypes (numbered 1-11) have been identified so far \citep{lazniewski2018structural}.

HA or NA proteins contain antigenic sites that are recognised by the immune system and which inhibit flu infection. These antigenic sites can rapidly alter in order to escape recognition by the immune system. There is a process known as antigenic drift that generates new influenza A, B, and C strains that are not fully recognised by human immune systems, thereby contributing to seasonal influenza. When influenza A virus proteins undergo drastic changes on antigenic sites, this can cause antigenic shifts. Antigenic shifts may result from the reassortment of different viruses within one or more hosts, leading to the emergence of new viruses \citep{brockwell2009diversity}. In the past, several pandemics have been the result of extreme antigenic shifts in which most people were incapable of resisting the novel virus. As a result of recombination between animal viruses (swine and avian) and human viruses, four major influenza epidemics have emerged since 1900: Spanish flu (1918–1919), Asian flu (1957–1958), Hong Kong flu (1968–1969), and the 2009 flu pandemic (2009–2010).

The virus responsible for the Spanish flu was A/H1N1. An estimated 17 - 100 million people died in this pandemic and it was the deadliest in recorded history \citep{spreeuwenberg2018reassessing}, \citep{morens20071918}, \citep{johnson2002updating}. Despite the mystery surrounding the origins of Spanish flu \citep{antonovics20061918}, recent studies suggest it may have originated in birds or pigs \citep{taubenberger2005characterization}, \citep{worobey2014synchronized}, \citep{smith2009dating}. The virus adapted and kept playing a major role in flu epidemics until 1957, when major changes in HA and NA resulted in the novel A/H2N2 virus and the Asian flu pandemic. The Asian flu is associated with a higher rate of morbidity and mortality compared to the Hong Kong flu of 1968, as the Hong Kong flu was caused by A/H3N2, which only changed the HA antigen \citep{kilbourne2006influenza}. The Asian flu and the Hong Kong flu occurred as a result of reassortment between human and avian viruses. A/H1N1 was also responsible for the 2009 flu pandemic, however, that iteration involved a triple reassortment between human, avian, and swine viruses \citep{garten2009antigenic}. \citep{smith2009origins}.

It is known that influenza viruses can infect a number of hosts, including humans, birds, pigs, and horses. Birds are a major natural reservoir for the influenza A virus \citep{long2019host}, \citep{gorman1990evolution}, which can infect humans and pigs alike \citep{webster1992evolution}. Additionally, pigs are considered to be an intermediate host for influenza A viruses between humans and birds \citep{brown2001pig}. The re-assortment of viruses between different hosts can result in life-threatening risks for human populations, since the viruses don't require an intermediate host to propagate.

The transmission of influenza viruses can, therefore occur through animal-to-human (zoonosis) as well as animal-to-animal (enzootic) contact \citep{long2019host}. Zoonotic infections can either be dead-end transmissions or lead to a pandemic in the human population after accumulating enough adaptive mutations to sustain transmissions between people, which then regularly circulates as a seasonal influenza virus \citep{long2019host},\citep{taubenberger2010influenza}. The origin of each virus during a virus outbreak is difficult to determine because some viruses can cross species boundaries. It is thus possible to isolate swine-origin viruses from humans. However, the virus must be given sufficient time to complete the adaptive mutation and accumulation process \citep{long2019host}. Thus, early isolation of the original viral host can effectively prevent or control the spread of a viral outbreak. 

Most traditional methods are laboratory-based, such as the use of hemagglutination inhibition (HI) assays to subtype viruses. Laboratory-based methods are laborious and time-consuming. To save manpower and time, a variety of machine learning algorithms have been used to predict viral hosts, such as k-nearest neighbours (KNNs) \citep{sherif2017classification}, random forests (RFs) \citep{sherif2017classification}, artificial neural networks (ANNs) \citep{attaluri2010applying}, and decision trees (DTs) \citep{kargarfard2016novel}. Most previous studies selected balanced data sets manually \citep{sherif2017classification}, \citep{Attaluri2010IntegratedUO}, used small data sets \citep{attaluri2010applying}, encoded the sequence as sparse matrices \citep{attaluri2010applying}, \citep{mock2021vidhop} or incorporated feature extraction procedures \citep{yin2018computational}. Moreover, some deep learning techniques, such as convolutional neural networks (CNN), have been applied in this field, although only to avian and human viruses thus far \citep{scarafoni2019predicting}.

We used various machine learning algorithms (i.e., RF, RUSBoost, SVM, XGB, MLP, Transformer, and CNN) to determine the origin host of influenza viruses. To vectorize a sequence, both alignment-based and alignment-free approaches are used. With respect to alignment-based sequence representation, evolutionary features of viral sequences were extracted by PSI-BLAST and fed into four traditional machine learning models. A word embedding technique is used as an alignment-free approach to process sequences. The morphological structure of sequence data is similar to that of text data, which suggests that methods for encoding sequences in natural language processing (NLP) could also be applied to process viral sequence data, such as word embedding. Word embeddings are a widely used and powerful technique for processing sequential data. By learning the meanings of words, similar words will have similar embeddings. In this study, instead of evaluating models only at a higher classification level, we divided avian classes and evaluated models at a lower classification level \citep{sherif2017classification}, \citep{attaluri2010applying},\citep{kargarfard2016novel}, \citep{Attaluri2010IntegratedUO}, \citep{mock2021vidhop}, \citep{xu2017predicting}, \citep{yin2018computational}. In addition, we discussed the impact of incomplete virus sequences on model performance. The results show that the Transformer neural network outperforms other models in most scenarios.

\section{Data}
In this study, two data sets were used. The primary difference between the two sets is that data set 2 includes both complete and incomplete HA protein sequences. Therefore, data set 2 is completely unforeseeable for models and noisier than data set 1. We used data set 2 as an additional testing set to test the performance of the pre-trained model further. The per-trained models were trained and validated by data set 1. 

\subsection{Data Set 1}
Data set 1 includes the complete Influenza A virus (IAV) hemagglutinin (HA) protein sequences isolated from avian, swine, and human samples in the GISAID \citep{gisaid} database (status 2020-09-25). Only HA protein sequences are used, as HA is the most dominant protein for immunity response and helps the virus bind to target hosts \citep{earn2002ecology}.  To maintain the quality of the data, we further removed sequences that are either redundant, multi-label, or contain amino acids X, B, and Z \citep{worthsone}. Therefore, a total of 59,785 sequences from the original set have been selected.  Only data set 1 was subjected to nested cross-validation (CV), as described in Section \ref{cv}.

\subsection{Data Set 2}
Data set 2 is an additional testing set. It includes the IAV HA protein sequences collected from 2020-09-26 to 2022-05-05 in the GISAID \citep{gisaid} database. Those sequences that appear in both data sets 1 and 2 are removed from data set 2, so the data sets are mutually exclusive. In addition, any redundant or multi-label sequences are also removed. Therefore, the sequences in data set 2 are also unique and have an unambiguous isolated host. The final set consists of 3,686 sequences. 

\begin{figure}[]
\setlength{\abovecaptionskip}{0.cm}
\centering
\includegraphics[width = \linewidth]{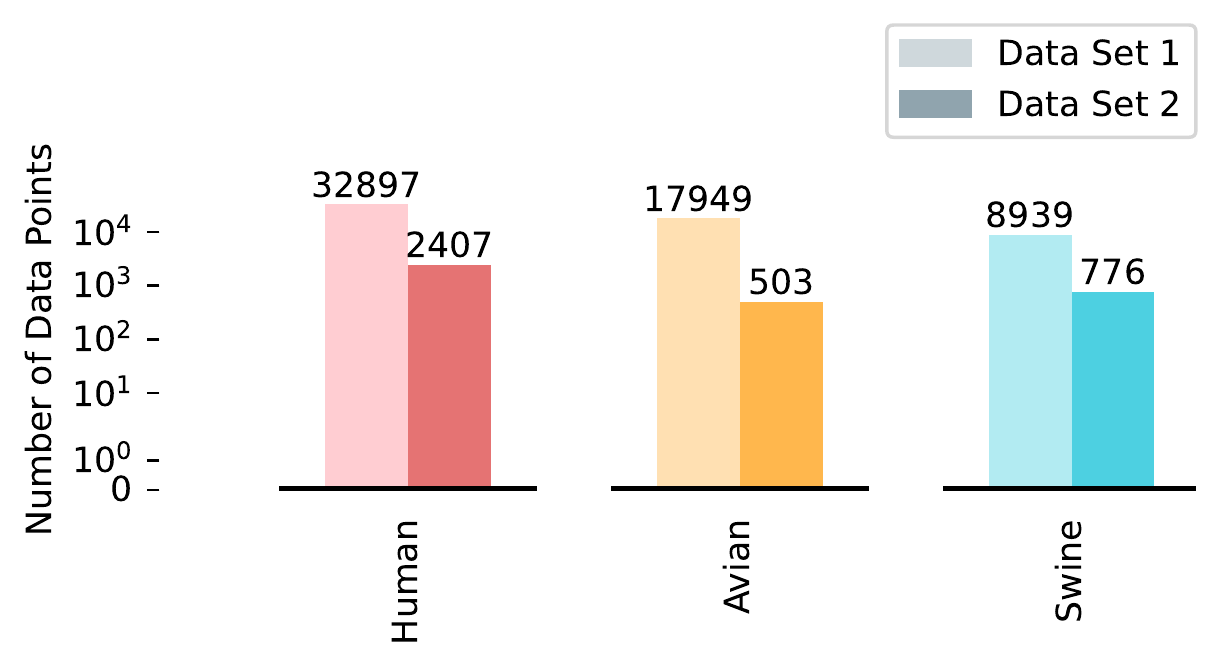}
\caption{Data distribution (higher taxonomic level)}
\label{fig_dataset_h}
\end{figure}

\begin{figure*}[]
\setlength{\abovecaptionskip}{0.cm}
\centering
\includegraphics[width = \linewidth]{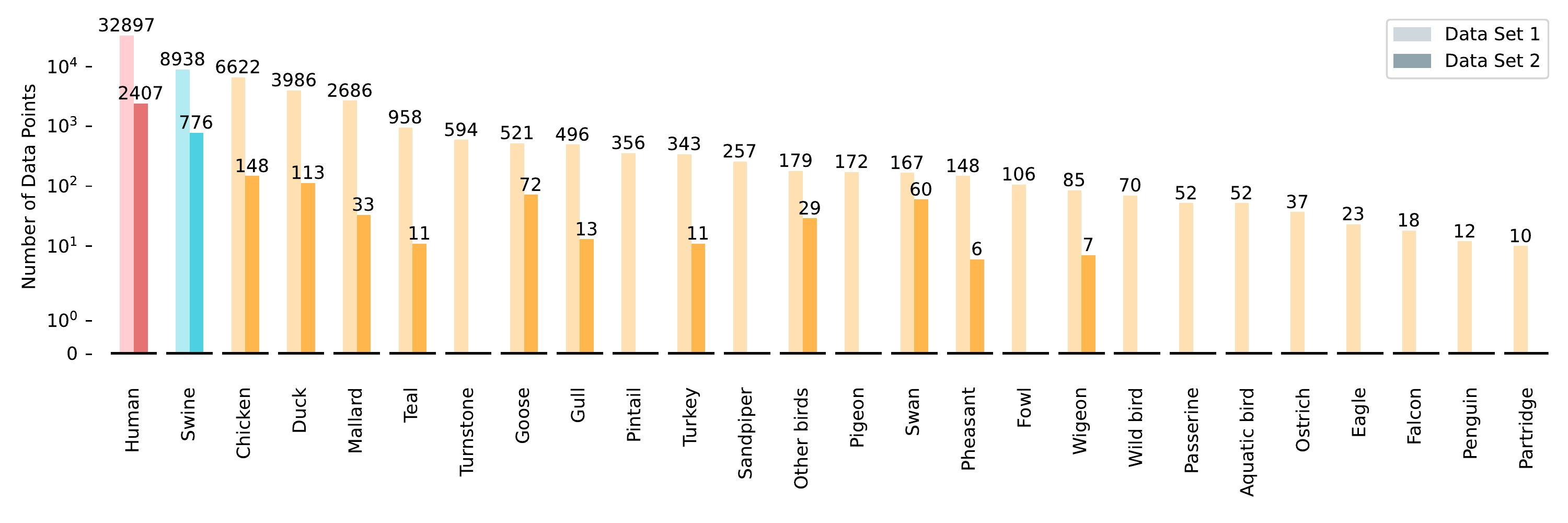}
\caption{Data distribution (lower taxonomic level)}
\label{fig_dataset_l}
\end{figure*}

The change in the amount of data before and after filtering appears in Table~\ref{tab_data}. Fig.~\ref{fig_dataset_h} and Fig.~\ref{fig_dataset_l} show the data distribution of different taxonomic levels. In terms of the performance of the model at a higher taxonomic level, we only consider three classes: human, avian, and swine, whereas at a lower classification level, avian data is divided further and results in 26 classes. When the number of classes increases, models face more challenges. 

\begin{table*}[]
\centering
\caption{Number of data before and after selection}
\ra{1.3}
\begin{threeparttable}
\begin{tabular}{ccccccccc}
\toprule
\textbf{Data Sets} & \textbf{Original \#} & \textbf{Selected \#}  & \textbf{Only Complete?} & \textbf{NR\tnote{a} ?} & \textbf{No Multi-label?} & \textbf{No X, B, Z?} & \textbf{Purposes\tnote{b}} \\
\midrule
1                  & 180,833              & 59,785                      &\checkmark                       & \checkmark                       & \checkmark                            &    \checkmark  & Train, Val and Test          \\
2                  & 13,798                & 3,686               &                         & \checkmark                       & \checkmark                            &                      & Test       \\
\bottomrule       
\end{tabular}
\begin{tablenotes}
    \item[a] NR: non-redundant
    \item[b] The purpose of the data set: training, validation or testing.
  \end{tablenotes}
 \end{threeparttable}
\label{tab_data}
\end{table*}

\section{Sequence Representations}
\subsection{Position-Specific Scoring Matrix-Based Representations}
\subsubsection{Position-Specific Scoring Matrix}
One of the most commonly used methods for extracting evolutionary information from protein sequences is the position-specific scoring matrix (PSSM) \citep{altschul1997gapped}, which can be generated using Position-Specific Iterated BLAST program (PSI-BLAST) \citep{altschul1998iterated}. 

As protein sequences typically contain 20 different types of amino acids (\emph{A}, \emph{R},  ... \emph{V}), a PSSM is a $L\times20$ matrix for a query protein sequence with $L$ length. The PSSM for the sequence a $a_1a_2\ldots a_L$ can be expressed as follows:

\vspace{-0.3 cm}
\begin{equation}\label{equ:pssm}
\mbox{\small\( %
{\rm PSSM}_{original}=\left(\begin{matrix}&A&R&\ldots&V\\a_1&p_{1,1}&p_{1,2}&\ldots&p_{1,20}\\a_2&p_{2,1}&p_{2,2}&\ldots&p_{2,20}\\\cdots&\cdots&\cdots&\cdots&\cdots\\a_L&p_{L,1}&p_{L,2}&\ldots&p_{L,20}\\\end{matrix}\right),
\)} %
\end{equation}

where $p_{i,j}$ is the score of the amino acid $a_i$ that mutates to $a_j$. It can also be interpreted as a probability of mutation in the range of [0,1] using the sigmoid function:

\vspace{-0.3 cm}
\begin{equation}
p_{i,j}=\frac{1}{\left(1+e^{-p_{i,j}}\right)},i=1,2,\ldots,L;j=1,2,\ldots20 ,
\end{equation}

We ran PSI-BLAST \citep{blastexe} iteratively with default parameters (E-value = 0.001, number of iterations = 3) on a non-redundant (nr) database. PSSMs cannot be fed directly into classic machine learning models due to their variable size. To overcome this hindrance, we propose three sequence encoding schemes based on PSSM. In order to reduce the complexity of proteins and unnecessary computations, we first introduce a residue grouping rule.

\subsubsection{Residue Grouping Rule}
As amino acids have similar properties in proteins, they can be classified into 10 groups \citep{li2003reduction}: G1 (F, \textbf{Y}, W), G2 (M, \textbf{L}), G3 (I, \textbf{V}), G4 (A, T, \textbf{S}), G5 (\textbf{N}, H), G6 (Q, \textbf{E}, D), G7 (R, \textbf{K}), G8 (C), G9 (G) and G10 (P).  A grouped-PSSM (GPSSM) with $L \times 10$ dimensions can be created by applying residue grouping rules to the columns of the original PSSM:

\vspace{-0.3 cm}
\begin{equation}\label{equ:gpssm}
\mbox{\small\( %
{\rm PSSM}_G=\left(\begin{matrix}&G_1&G_2&\ldots&G_{10}\\a_1&g_{1,1}&g_{1,2}&\ldots&g_{1,10}\\a_2&g_{2,1}&g_{2,2}&\ldots&g_{2,10}\\\cdots&\cdots&\cdots&\cdots&\cdots\\a_L&g_{L,1}&g_{L,2}&\ldots&g_{L,10}\\\end{matrix}\right),
\)} %
\end{equation}

where 

\vspace{-0.3 cm}
\begin{equation}
g_{i,j}=\frac{\sum p_{i,G_j}}{\left|G_j\right|},
\end{equation}

The GPSSM is produced based on the original PSSM~\eqref{equ:pssm}, thence $\sum p_{i,G_j}$ represents the score of an amino acid $a_i$ that is mutated to an amino acid belonging to group $j$. $L$ is the length of sequences; $i = 1, 2, \ldots, L$; $\left|G_j\right|$ is the number of amino acid types in group $j$. The GPSSM~\eqref{equ:gpssm} was used to derive the following proposed feature sets: EG-PSSM, GDPC-PSSM, and ER-PSSM.

\subsubsection{EG-PSSM}
The length of the input sequence as well as the original PSSM~\eqref{equ:pssm} and GPSSM~\eqref{equ:gpssm} may vary. This means that they cannot be directly used in many machine learning models. There is an intuitive method for overcoming this problem by applying the residue grouping rule across rows of the GPSSM~\eqref{equ:gpssm}. This will result in a matrix of $10\times10$. We reformat the matrix  by iterating through it row by row, moving from left to right within each row, thereby generating a $1\times100$ feature vector from a GPSSM~\eqref{equ:gpssm} before feeding it to classical machine learning models:

\vspace{-0.3 cm}
\begin{equation}
{\rm PSSM}_{EG}=\left(\begin{matrix}E_{G_1,G_1}&E_{G_1,G_2}&\begin{matrix}\cdots&E_{G_{10},G_{10}}\\\end{matrix}\\\end{matrix}\right)^T,
\end{equation}

where 

\vspace{-0.3 cm}
\begin{equation}
E_{G_i,G_j}=\frac{\sum g_{G_i,G_j}}{\left|G_i\right|}, i,j = 1 \ldots 10,
\end{equation}
\subsubsection{GDPC-PSSM}
By using dipeptide compositions (DPCs) \citep{saravanan2015harnessing}, we are able to determine amino acid composition information and partial local-order information in protein sequences. DPC acts directly on raw sequence data and generates a 400-dimensional feature vector for each sequence, but it can also be extended to PSSM \citep{liu2010prediction}. Therefore, each $L \times 10$ GPSSM~\eqref{equ:gpssm} can be rewritten as a $10\times10$ matrix by grouped dipeptide composition encoding. This $10\times10$ matrix can then be reshaped as a 100-dimensional feature vector:

\vspace{-0.3 cm}
\begin{equation}
{\rm PSSM}_{GDPC}=\left(\begin{matrix}D_{1,1}&D_{1,2}&\begin{matrix}\cdots&D_{10,10}\\\end{matrix}\\\end{matrix}\right)^T,
\end{equation}

where 

\vspace{-0.3 cm}
\begin{equation}
D_{i,j}=\frac{1}{L-1}\sum_{k=1}^{L-1}{g_{k,i}\times g_{k+1,j}}\ i,j=1,2,\ldots,10,
\end{equation}

Each $g_{k,i}$ is the value of row $k$ and column $i$ in the GPSSM~\eqref{equ:gpssm}.

\subsubsection{ER-PSSM}
The third proposed sequence representation is adapted from RPSSM \citep{ding2014protein}, which computes the pseudo-composition of the dipeptide in sequences. As with GDPC-PSSM, RPSSM also extracts partial local sequence order information from sequences. RPSSMs only compute the pseudo-composition of any two adjacent amino acids. We extended the computation of RPSSM for any two amino acids $a_ka_{k+t}$ with gap $t$ in sequences and extracted a $91\times10$ matrix per sequence, the matrix can be reformatted as a $1\times910$ feature vector: 

\vspace{-0.3 cm}
\begin{equation}
\resizebox{.9\hsize}{!}{${\rm PSSM}_{ER}=\left(\begin{matrix}M_{1,1,1}&M_{1,2,1}&\begin{matrix}\cdots&\begin{matrix}M_{10,10,9}&\begin{matrix}T_1&\begin{matrix}\cdots&T_{10}\\\end{matrix}\\\end{matrix}\\\end{matrix}\ \\\end{matrix}\\\end{matrix}\right)^T $},
\end{equation}

where

\vspace{-0.3 cm}
\begin{equation}
\begin{split}
M_{i,j,t}=\frac{1}{L-t}\sum_{k=1}^{L-t}\frac{\left(g_{k,i}-g_{k+t,j}\right)^2}{2},\ \\
i,j=1,2,\ldots,10;t=1,2,\ldots,9
\end{split}
\end{equation}

and 

\begin{equation}
T_i=\frac{1}{L}\sum_{k=1}^{L}{\left(g_{k,i}-\bar{G_i}\right)^2,\ \ \ }i,j=1,2,\ldots,10.
\end{equation}

$\bar{G_i}$ is the average of values of GPSSM~\eqref{equ:gpssm} in column $i$, $T_i$ computes the average pseudo-composition of all the amino acids in the protein sequence corresponding to column $i$ in GPSSM~\eqref{equ:gpssm}.

\subsection{Learning Representations}
Traditional machine learning algorithms require manual preprocessing and feature extraction to extract representative features from each protein sequence. A priori knowledge is required to select suitable features during the feature extraction process. Conversely, deep learning algorithms can learn the implicit representations of protein sequences directly. In this study, a powerful vectorisation scheme in natural language processing (NLP), word embedding, is applied to map protein sequences into numerical vectors.

\subsubsection{Overlapping N-grams}
\begin{figure}[h]
\setlength{\abovecaptionskip}{0.cm}
\centering
\includegraphics[scale = 0.95]{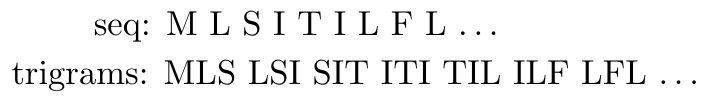}
\caption{Example of overlapping trigrams: the protein sequence MLSITILFL can be converted into a protein "sentence" containing 7 protein "words" MLS, LSI, SIT, ITI, TIL, ILF and LFL.}
\label{fig_3gram}
\end{figure}

\begin{figure}[h]
\setlength{\abovecaptionskip}{0.cm}
\centering
\subfloat[Avian]{%
  \includegraphics[width=0.16\textwidth]{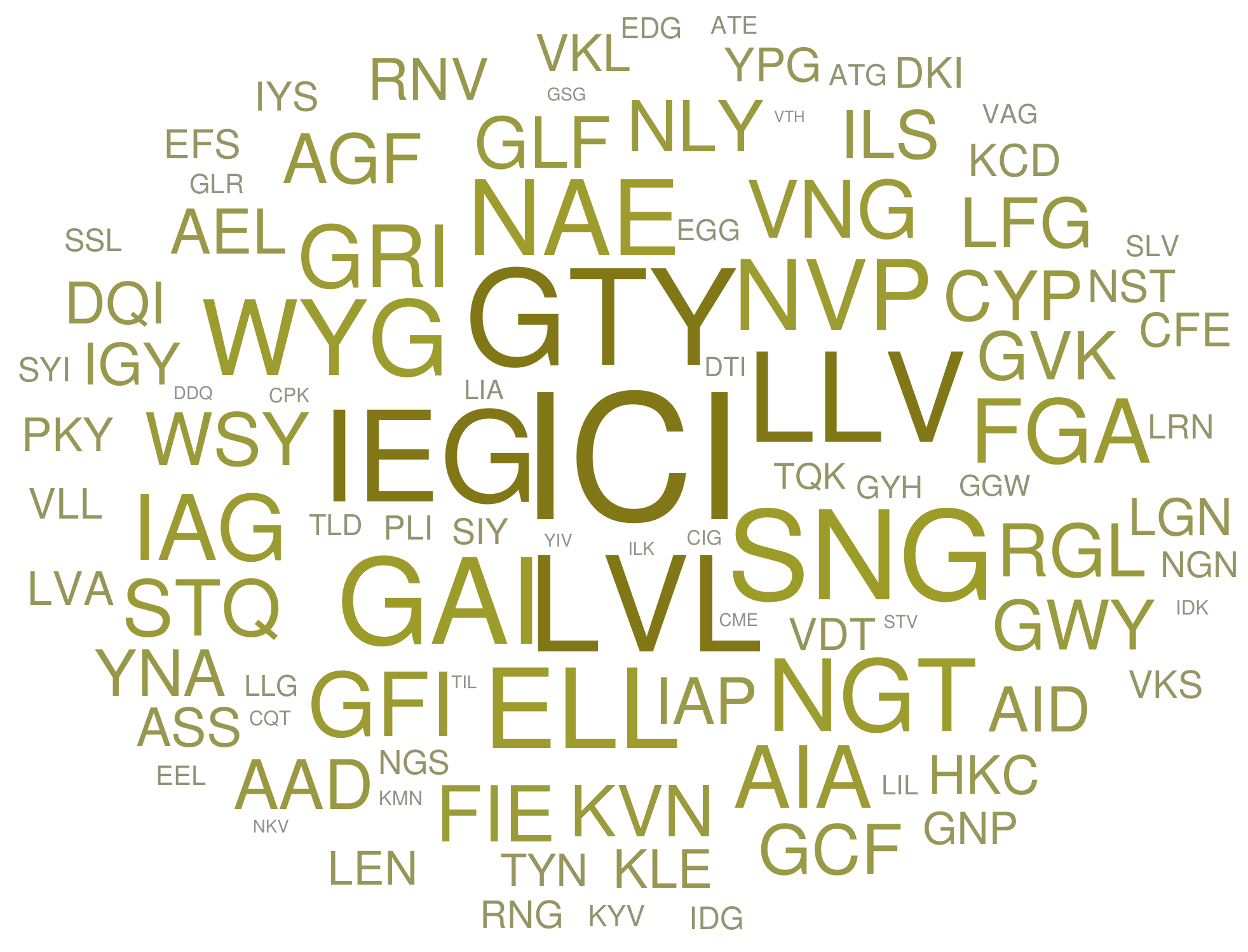}
}
\subfloat[Human]{%
  \includegraphics[width=0.16\textwidth]{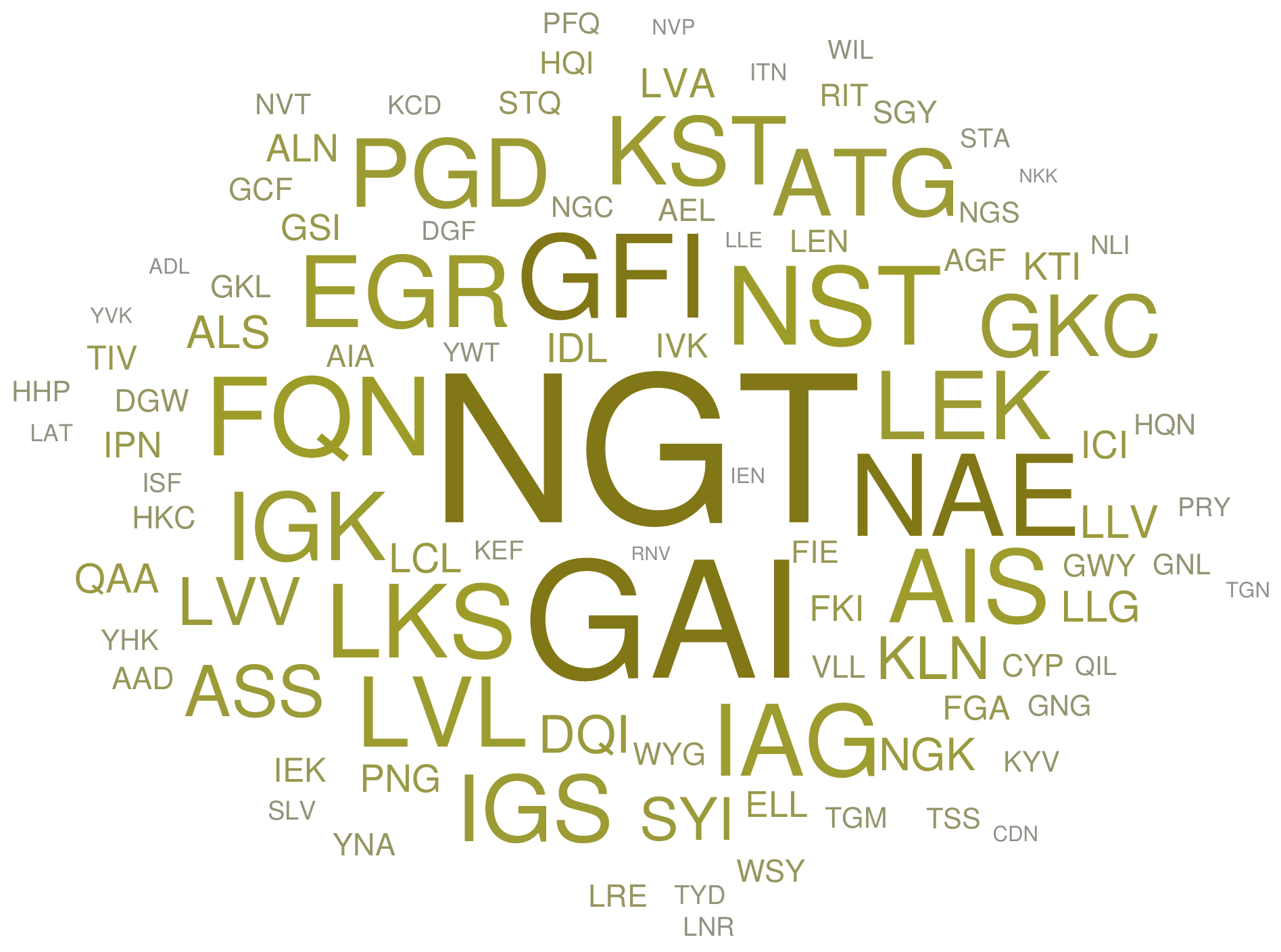}
}
\subfloat[Swine]{%
  \includegraphics[width=0.16\textwidth]{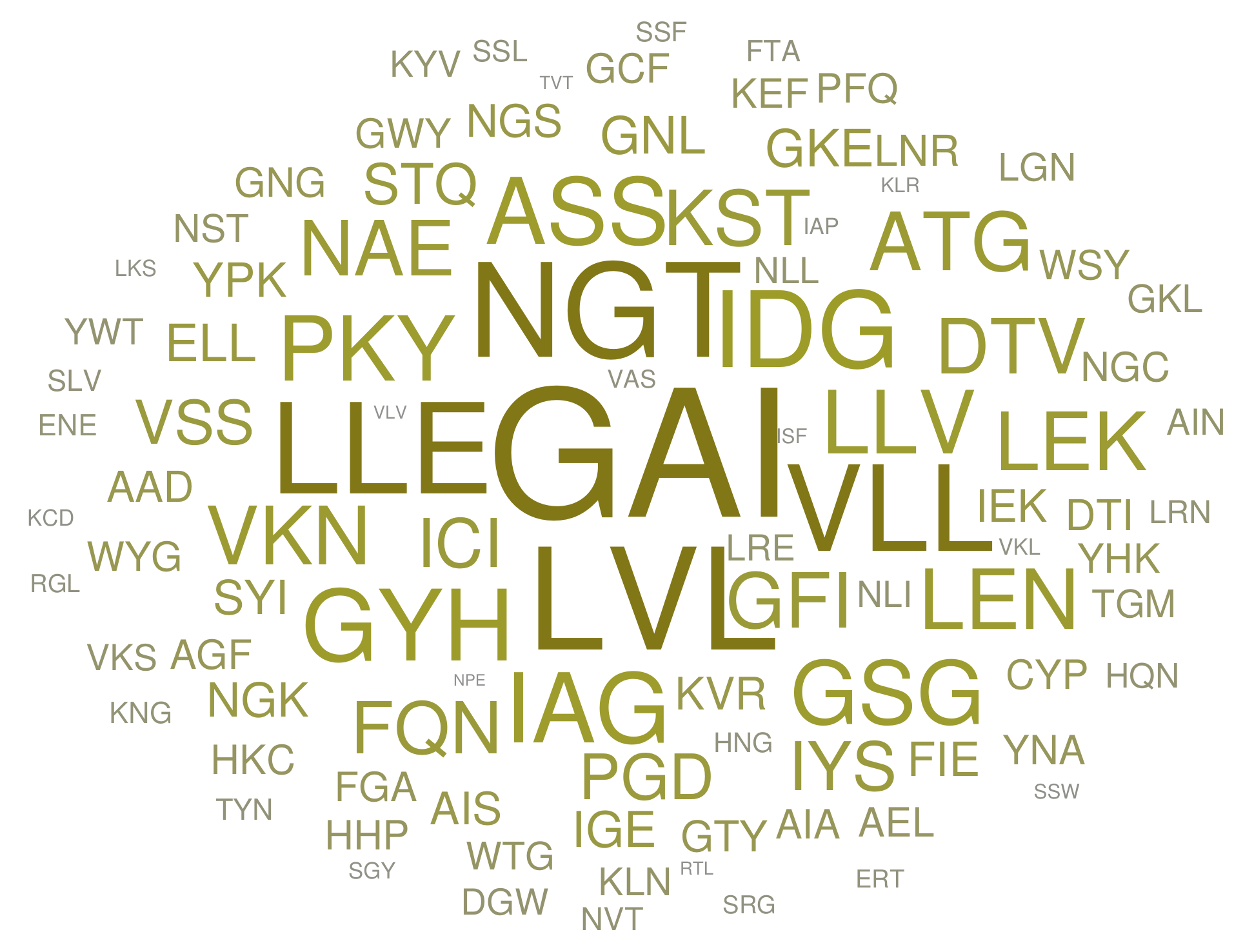}
}
\caption{Word clouds of trigrams for each class, generated by MATLAB\textsuperscript{®}}
\label{fig_wordcloud}
\end{figure}

Protein sequences are morphologically similar to text sentences, except that text is composed of words whereas amino acid letters comprise a protein sequence. In order to transform a protein sequence into a protein "sentence", we split the sequence into overlapping n-grams (n varies from 3 to 5). An n-gram is a protein "word" composed of successive n amino acids. Fig.~\ref{fig_3gram} is an example of overlapping 3-grams for a protein sequence and Fig.~\ref{fig_wordcloud} shows the word clouds of trigrams for human, avian and swine.

\subsubsection{Word Embedding}
Word embedding overcomes the drawbacks of one-hot encoding. It cannot only produce dense vectors but also capture the relationship between similar words. It is capable of producing dense vectors as well as capturing relationships between words. The idea behind word embedding is to map words to an embedding space, where words with similar meanings are closer together, and hence have similar embeddings. Word2Vec \citep{mikolov2013efficient} is a popular implementation of word embedding, but it does not include domain-specific words. Thus, we generated a custom word embedding from the training set and mapped the n-grams of each sequence to the embedding vectors. An n-gram is represented as a vector of size $N$, and a protein sequence is represented as a $L \times N$, where $L$ is the length of the sequence (number of n-grams in the sequence), and $N$ is the embedding dimension. 

To unify the dimensions of matrices, we use left-padding on the sequences to match the longest sequence length. Therefore, most sequence information will be retained, however, more noise will be introduced to the shortest sequence.

\section{Machine Learning Algorithms\label{ml}}
\subsection{RUSBoost}
Class imbalance can be addressed using data sampling and boosting algorithms. Oversampling (enriching minority classes) and undersampling (decreasing majority classes) are common methods of data sampling. Random undersampling boosting (RUSBoost)  \citep{seiffert2008rusboost} is an algorithm that uses undersampling together with boosting techniques. RUSBoost is computationally cheaper and more efficient than other oversampling methods, such as SMOTEBoost \citep{chawla2003smoteboost}.

\subsection{Extreme Gradient Boosting}
\begin{sloppypar}
Extreme Gradient Boosting (XGBoost) \citep{10.1145/2939672.2939785} implements gradient boosting algorithms in a scalable and efficient fashion. Gradient boosting algorithms are similar to AdaBoost, except that gradient boosting algorithms use gradient descent to optimise the derivable loss function when adding new models. XGBoost can help in handling tough problems in data science, such as solving missing values and sparse data automatically. One of the biggest advantages of XGBoost is that it provides parallel training to speed up the training process and can handle large datasets.
\end{sloppypar}

\subsection{Random Forest}
The bagging algorithm reduces the variance of the model, while decision trees have higher variance and lower bias. Therefore, this model performs better when bagging and decision trees are combined (random forest). Random forests \citep{ho1995random} consider only a small portion of all features in each split. Each decision tree is thus more random. Contrary to boosting-based ensembles, bag-based ensembles tend to construct deep trees, which means bag-based ensembles are more complicated. In this respect, bag-based ensembles may, in some instances, require more training time than boosting-based ensembles, but omit the validation process to estimate generalisation performance.

\subsection{Support Vector Machine}
Support Vector Machine (SVM) is one of the most commonly used supervised learning algorithms \citep{gove2012machine}. In addition to being able to classify linearly separable data, SVM can also classify non-linearly separable data by introducing a kernel trick. The kernel trick allows SVMs to successfully handle high-dimensional (even infinite-dimensional) data by mapping lower-dimensional data into higher-dimensional data without explicitly transforming it. We construct the multi-class SVM by using the Gaussian kernel function as the kernel function.

\subsection{Multi-Layer Perceptron}
\begin{figure}[h]
\setlength{\abovecaptionskip}{0.cm}
\centering
\includegraphics[]{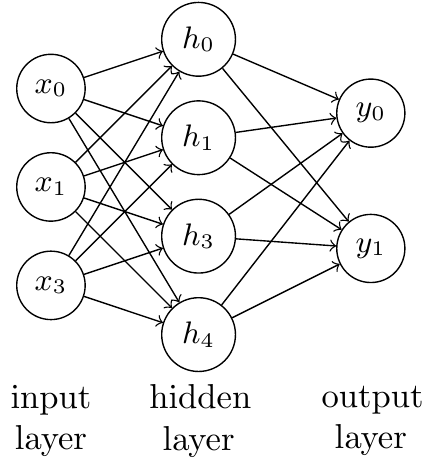}
\caption{Example of a fully connected MLP architecture}
\label{fig_mlp}
\end{figure}

Multi-Layer Perceptron (MLP) is a feed-forward neural network. It compensates for the limitation of a single-layer perceptron that cannot process linearly separable problems \citep{minsky2017perceptrons}.  A simple MLP usually consists of three layers: an input layer, a hidden layer, and an output layer. Fig.~\ref{fig_mlp} shows an example of a three-layer fully connected MLP, where the input layer, hidden layer, and output layer have three, four, and two neurons, respectively. The number of neurons in the input layer equals the number of features in the input data, and the number of classes in the data set determines the number of neurons in the output layer.

\subsection{Convolutional Neural Network}
Convolutional neural networks (CNNs) are widely used in a variety of fields, including facial recognition, object recognition, and autonomous vehicles. CNNs were initially trained on images and expanded to take in other types of data, such as time series, text, and audio data. As opposed to traditional machine learning, CNNs learn data features in each hidden layer. CNNs also differ from standard fully connected neural networks (Fig.~\ref{fig_mlp}) in that they sparsely connect layers and reduce the number of parameters that must be learned. CNNs typically include three layers: the convolutional layer for learning spatial features, the activation layer for activating features, and the pooling layer for downsampling the features.

\begin{figure*}[h]
\setlength{\abovecaptionskip}{0.cm}
\centering
\includegraphics[width = \linewidth]{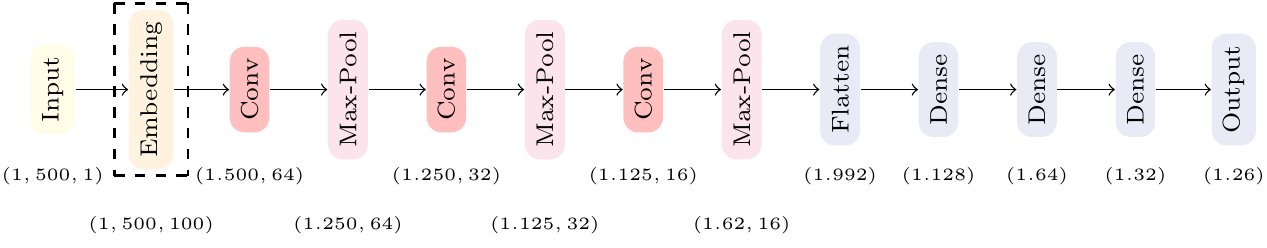}
\caption{Example of the CNN architecture with an embedding layer: here for simplicity, we assume that the longest length of the sequence is 500, the dimension of embedding is 100, and the number of filters for the first, second, and third convolutional layers is 64, 32, and 16, respectively. The layers are represented by coloured boxes: input layer (light yellow), embedding layer (light orange), convolutional layer (light red), max-pooling layer (light pink), and flatten/dense layer (light blue).}
\label{fig_cnn_architecture}
\end{figure*}

We designed a simple CNN for classifying protein sequences. The CNN in this study consists of one input layer (\emph{Input}), three convolution layers (\emph{Conv}), three max-pooling layers (\emph{Max-Pool}), one flattening layer (\emph{Flatten}), three dense layers with Rectified Linear Unit (ReLU) activation  (\emph{Dense})  and one dense layer with softmax activation (\emph{Output}). PSSM or tokenised virus sequences can be used as input data. The CNN also includes an embedding layer (\emph{Embedding}) when the input data are tokenised sequences, as shown in  Fig.~\ref{fig_cnn_architecture}. The hyperparameter settings for CNN can be seen in Table~\ref{tab_tunning}.

\subsection{Transformer}
Transformer neural network is an improvement on normal recurrent neural networks (RNNs) and performs better than many state-of-the-art models on the Winograd schemas language translation tasks \citep{ackerman2014can} according to Bilingual Evaluation Understudy (BLEU) scores \citep{vaswani2017attention}.  Typical RNNs are slow to train sequential data as they must process words in order resulting in a lack of parallelisation capability. Additionally, RNNs cannot handle long sequences very well due to vanishing and exploding gradients phenomena.

Contrary to RNNs, Transformer networks abandon recurrence and embrace self-attention mechanisms. The attention mechanism, as the name suggests, focuses on the parts of the input it deems important. The attention takes a query and key-value pairs as input. The specific attention mechanism used in the classic Transformer network is scaled dot-product attention, expressed in the following formula:\begin{equation}
Attention(Q, K, V) = softmax(\frac{QK^T}{\sqrt{d_k}})V
\end{equation}
where \(d_k\) is the dimension of the key, so \(\nicefrac{1}{\sqrt{d_k}}\) is the scaling factor. \(Q\), \(K\)and \(V\) denote for query vector, key vector and value vector, respectively. The softmax function converts the attention score to attention distribution. The core idea behind dot-product attention is that the dot product is higher between similar sequences than in dissimilar ones. 

One of the innovations of the Transformer is multi-head attention. The multi-head attention assembles multiple scaled dot-product attentions. Compared with single-head attention, multi-head attention avoids words that focus themselves too much and produces more robust results. Similar to bag-of-words, attention lacks sequence order information. The input embedding layer  (\emph{Input Embedding}) retrieves the meaning of words but not their positions. To compensate for this drawback, the original paper \citep{vaswani2017attention} adds positional encoding to input embedding, and in this paper, we added positional embedding (\emph{Positional Embedding}) for the same purpose. More details of the Transformer can be referred to \citep{vaswani2017attention}.

\begin{figure*}[h]
\setlength{\abovecaptionskip}{0.cm}
\centering
\includegraphics[width = 0.85\linewidth]{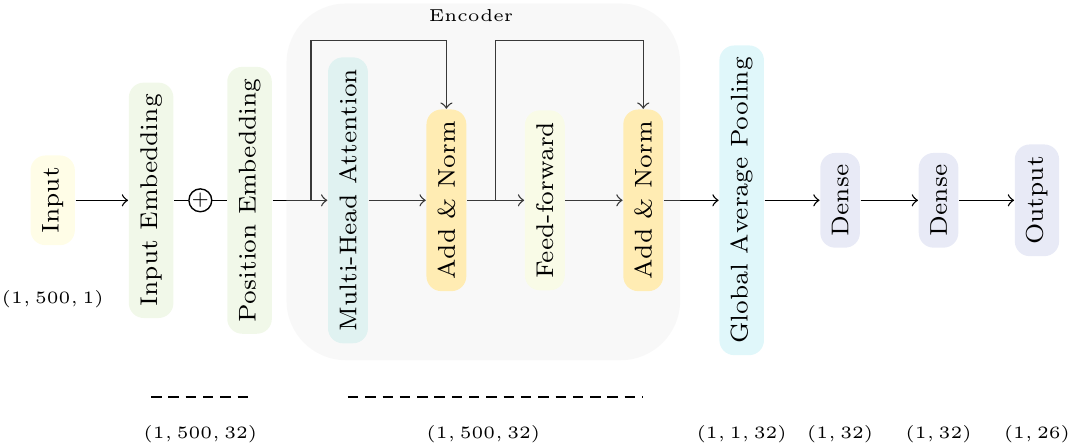}
\caption{Example of the Transformer architecture: here we assume that the number of heads is 3 and the dimension of embedding is 32.}
\label{fig_trans_architecture}
\end{figure*}

The sequence-sequence Transformer model usually includes an encoder block and a decoder block which take a sequence as input and output a sequence, such as the sequence-to-sequence model often used in language translation. In contrast, a sequence-to-vector model accepts a sequence as input and outputs a class label for that sequence; no decoder block is needed. We used a sequence-vector Transformer model, which accepts a sequence as input and outputs a class label for that sequence. Thus, the Transformer used in this study only has an encoder block. The Transformer architecture used in this study is shown in Fig.~\ref{fig_trans_architecture}. The number of heads (\emph{num\_heads}) ranges from 1 to 5, and the dimension of embedding (\emph{embed\_dim}) varies between 32, 64, and 128.

\section{Model Implementation and Evaluation \label{evaluation}}
\subsection{Model Implementation}
All models were evaluated using optimised parameters, as shown in Table~\ref{tab_tunning}. We used Bayesian optimisation to automatically adjust hyperparameters in 5 iterations. RF, SVM and MLP were implemented by the Scikit-learn \citep{scikit-learn}, RUSBoost was implemented by imblearn \citep{JMLR:v18:16-365}, XGBoost was implemented in XGBoost Python package \citep{10.1145/2939672.2939785}.  CNN and Transformer mode were implemented by Keras \citep{chollet2015keras}. The data is available at \url{https://github.com/dkdjb/IAV_Host_Prediction}.

\begin{table}[]
\caption{Hyperparameter settings}
\label{tab_tunning} 
\centering
\footnotesize
\ra{1.3}
\begin{tabular*}{\linewidth}{l@{\extracolsep{\fill}}@{}ll@{}}
\toprule
\multicolumn{1}{c}{\textbf{ Algorithms }} & \multicolumn{1}{c}{\textbf{Hyperparameters}} \\
 \midrule
SVM & \begin{tabular}[c]{@{}l@{}}C = 0.01, 0.1, 1, 10, 100\\ gamma = 0.001, 0.01, 0.1, 1\end{tabular} \\
\hdashline
RF & \begin{tabular}[c]{@{}l@{}}n\_estimators = 100, 200, 500, 1000, \\ 1500, 2000\\ max\_depth = 5, 10, 15, 20\end{tabular} \\
\hdashline
RUSBoost & \begin{tabular}[c]{@{}l@{}}n\_estimators = 50, 100, 200, 500, 1000, \\ 1500, 2000\\ learning\_rate = 0.001, 0.01, 0.1\end{tabular} \\
\hdashline
XGBoost & \begin{tabular}[c]{@{}l@{}}max\_depth = 5, 10, 15, 20\\ eta = 0.001, 0.01, 0.05\\ colsample\_bytree = 0.1, 0.3, 0.5, 0.8, 1\end{tabular} \\
\hdashline
MLP & \begin{tabular}[c]{@{}l@{}}alpha = 0.001, 0.01, 0.05\\ max\_iter = 500\\ learning\_rate\_init = 0.001, 0.01, 0.05\end{tabular} \\
\hdashline
CNN & \begin{tabular}[c]{@{}l@{}}num\_filters = 64, 128,  256\\ learning\_rate = 0.01, 0.05, 0.001, 0.0001\\ batch\_size = 128\\ epochs = 300 \\ kernel\_size = 3 \end{tabular} \\
\hdashline
Transformer & \begin{tabular}[c]{@{}l@{}}embed\_dim = 32, 64, 128\\ num\_heads = 1, 2, 3, 4, 5\\ batch\_size = 128\\ epochs = 300\end{tabular} \\
\bottomrule
\end{tabular*}
\end{table}

\subsection{Cross-Validation\label{cv}}
\begin{figure}[h]
\setlength{\abovecaptionskip}{0.cm}
\centering
\includegraphics[width = \linewidth]{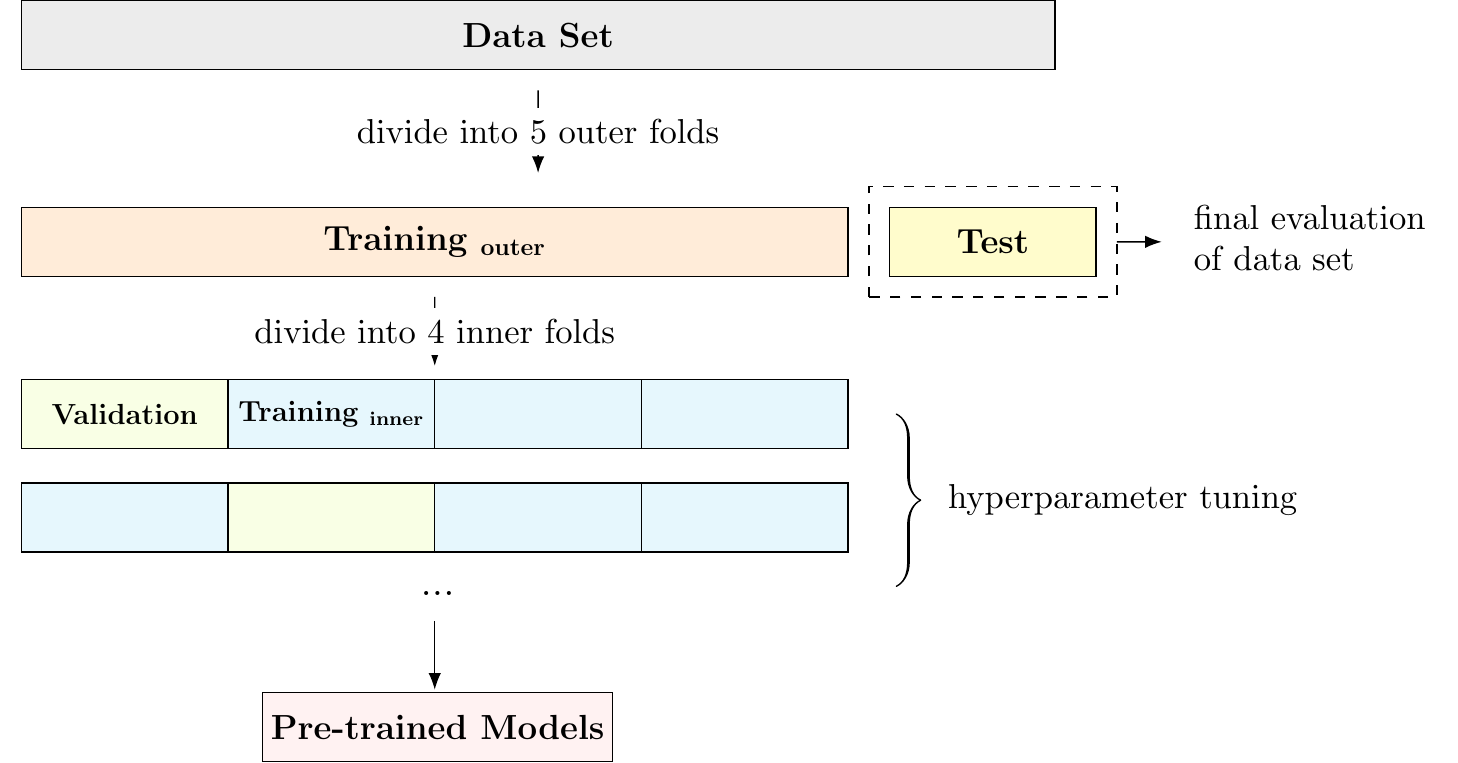}
\caption{Example of the nested $K$-fold CV ($k_{outer}=5$ and $k_{inner}=4$): we used models trained during nested cross-validation (i.e., pre-trained models) to predict unseen data. In this study, data set 2 and a proportion of data set 1 served as unseen data for pre-trained models.}
\label{fig_cv}
\end{figure}

Stratified $K$-fold cross-validation (CV) was used to evaluate models. The class ratio of the training set was almost the same as that of the test set. The generalisation performance of models was only evaluated on the test set (unseen data to the model). Nested $K$-fold CV adds the outer $K$-fold CV for final evaluation to reduce bias when it comes to hyperparameters optimisation or model selection \citep{cawley2010over}. Therefore, a nested $K$-fold CV will take advantage of the full diversity of the data set and ensure that all data will be tested. The example of stratified nested $K$-fold CV is shown in Fig.~\ref{fig_cv}.

In this study, we chose $k_{outer}=5$ and $k_{inner}=4$.  Each outer fold includes approximately 20\% of the data from data set 1, serving as the test set  in each iteration. The remaining 80\% of the data forms the $training_{outer}$ set, which is further split into four inner folds for model training and validation. Specifically, within the inner loop, one of the inner folds (representing 25\% of the $training_{outer}$ set, which is 20\% of the total dataset) is used for validation. The other three (representing 75\% of the $training_{outer}$ set, which is 60\% of the total dataset) are used for training during each inner loop iteration.
\subsection{Evaluation Metrics}
Evaluation measurements used in the study include \emph{$F_1$}-score, Area Under Precision-Recall Curve (\emph{AUCPR}) and Matthews’s correlation coefficient (\emph{MCC}). The equations of  \emph{$F_1$}-score and MCC for each class are defined as follows:

\vspace{-0.2 cm}
\begin{equation}
F_{1_i}=\ 2\ \cdot\frac{\rm Precision_i \cdot \rm Recall_i}{\rm Precision_i+ \rm Recall_i},
\end{equation}
\vspace{-0.2 cm}
\begin{equation}
\resizebox{.85\hsize}{!}{${MCC}_i=\frac{TP_i\times T N_i-FP_i\times F N_i}{\sqrt{\left(TP_i+FP_i\right)\left(TP_i+FN_i\right)\left(TN_i+FP_i\right)\left(TN_i+FN_i\right)}}$},
\end{equation}

where ${\rm Recall}_i=\nicefrac{TP_i}{(TP_i+FN_i)}$, ${\rm Precision}_i=\nicefrac{TP_i}{(TP_i+FP_i)}$, $i=1,2,\ldots,\ N$, $N$ is the number of classes;  $TP$ (True Positive) and $TN$ (True Negative) represent the number of data correctly predicted, $FP$ is the number of negative data misclassified as positive, and $FN$ counts the number of positive data incorrectly predicted as negative. For multi-class classification, the one-vs-all strategy is applied to produce \emph{$F_1$}-score, for each class. 

The overall \emph{$F_1$}-score (\emph{O$F_1$}) and overall AUCRP (\textit{OAUCRP}) are the micro-average of  the corresponding metric to each class, and the overall MCC (\textit{OMCC}) is defined as follows:

\vspace{-0.2 cm}
\begin{equation}
{\displaystyle {{\rm Overall} \  MCC}={\frac {c \cdot s- {\sum_{i}^{N}{p_i}}\cdot{{t_i}}}{{\sqrt {s^{2}-{\sum_{i}^{N}{p_i^2}} }}\cdot{\sqrt {s^{2}-{\sum_{i}^{N}{t_i^2}}}}}}}
\end{equation}

where $t_{i}$ is the number of times that class $i$ truly occurred, $p_{i}$ is the number of times class $i$ was predicted, $c$ is the total number of correctly predicted data, $s$ is the total number of data items.

In Section \ref{results}, we also used the mean score to present the overall performance of each model. The mean score is defined as the mean of  OAUCPR, O$F_1$ and OMCC:
\begin{equation}
{\rm Mean\ Score} = average(OAUCPR, OF_{1}, OMCC)
\end{equation}

Regarding the models' 	performance in each class, we only show the results of AUCPR as AUCPR is recommended \citep{branco2016survey} for evaluating classifiers when data is highly imbalanced. 

\section{Results\label{results}}
\subsection{Performance at Different Taxonomic Levels}
We evaluated all models at different taxonomic levels. The higher taxonomic level represents only three classes: avian, human, and swine, while at a lower taxonomic level, the avian class was further subdivided, resulting in a total of 26 classes. The performance of sequence representations and machine learning algorithms across data set 1 are provided in Section \ref{comp_seqs} and \ref{comp_ml}. Section \ref{overall} presents the overall results of each model. 

\subsubsection{Comparison of Sequence Representations\label{comp_seqs}}

In order to encode protein sequences, we used two kinds of representation: sequence alignment-free (word embedding) and sequence alignment-based (PSSM-based representations). Table~\ref{tab_comparison_seqs} presents a comparison of sequence representation with metric scores averaged across machine learning algorithms. At lower and higher taxonomic levels, 3-grams word embeddings reach a mean score of 87.73\% and 97.94\%, respectively. In contrast, PSSM-based representations have wider variability due to their instability and poor performance on RUSBoost. When comparing PSSM-based representations, EG-PSSM has a relatively low deviation and a higher mean score.

At the lower taxonomic level, data becomes more skewed, with mean score drops for all sequence representations ranging from 10\% to 15\%. ER-PSSM is the most affected, with a mean score drop of 14.85\%, while 3-grams word embeddings are the least affected with a mean score drop of 10.21\%.

\begin{table*}[]
\caption{Comparison of sequence representations on data set 1.}
\label{tab_comparison_seqs}
\centering
\ra{1.3}
\begin{threeparttable}
\resizebox{\textwidth}{!}{
\begin{tabular}{cccccccccc}
\toprule
\textbf{Representations}  & \multicolumn{4}{c}{\textbf{Higher Classification Level}}                                     && \multicolumn{4}{c}{\textbf{Lower Classification Level}}                               \\ \cline{2-5} 
\cline{7-10}
& \textbf{Mean Score (\%)} & \textbf{AUCPR (\%)}   & \textbf{F1(\%)}   & \textbf{MCC (\%)}  && \textbf{Mean Score (\%)} & \textbf{AUCPR (\%)}   & \textbf{F1 (\%)}   & \textbf{MCC (\%)}  \\
\midrule
\textbf{WE\tnote{*} (2-grams)}   & 97.83 (0.19)        & 99.50 (0.07)  & 97.78 (0.19) & 96.21 (0.32)  && 87.42 (0.74) & 94.64 (0.42) & 87.18 (0.71) & 80.45 (1.09) \\
\textbf{WE (3-grams)}   & \textbf{97.94 (0.16)}        & \textbf{99.51 (0.05)}  & \textbf{97.90 (0.16)} &\textbf{ 96.41 (0.29)}  && \textbf{87.73 (0.56)} & \textbf{94.76 (0.35)} & \textbf{87.52 (0.55)} &\textbf{ 80.92 (0.81)} \\
\textbf{WE (4-grams)}   & 97.54 (0.36)        & 99.41 (0.09) & 97.48 (0.36) & 95.72 (0.61) && 87.21 (0.47) & 94.41 (0.35) & 87.02 (0.43) & 80.22 (0.65) \\
\textbf{WE (5-grams)}   & 97.32 (0.85)        & 99.36 (0.22) & 97.26 (0.87) & 95.34 (1.46) && 86.56 (1.33) & 93.97 (0.94) & 86.41 (1.24) & 79.30 (1.80) \\
\textbf{EG-PSSM}   & 92.52 (8.02)        & 96.02 (5.93) & 92.93 (7.33) & 88.62 (10.85) && 79.30 (10.81) & 87.78 (10.06) & 79.97 (9.54) & 70.16 (12.88) \\
\textbf{ER-PSSM}   & 89.77 (13.81)       & 94.15 (9.40) & 90.20 (13.72) & 84.95 (18.38) && 75.42 (15.29) & 85.42 (13.63) & 76.15 (14.18) & 64.68 (18.99) \\
\textbf{GDPC-PSSM} & 85.13 (21.06)       & 90.96 (14.46) & 85.50 (20.94) & 78.92 (28.00) && 70.28 (19.79) & 82.15 (18.08) & 72.90 (16.54) & 55.77 (30.31)
\\
\bottomrule
\end{tabular}
}
\begin{tablenotes}
    \item[*] {\small WE: Word Embedding}
  \end{tablenotes}
\end{threeparttable}
\end{table*}

\subsubsection{Comparison of Machine Learning Algorithms \label{comp_ml}}

Table~\ref{tab_comparison_ml} represents the comparison of machine learning algorithms with the averaged metric scores across sequence representations. RUSBoost performs the worst at both classification levels, but it is the only algorithm with narrower variability at higher classification levels than at lower classification levels. Contrary to RUSBoost, SVM has a larger deviation when the data is more skewed and the class increases. Therefore, the performance of RUSBoost and SVM is most dependent on the sequence representation, but RUSBoost is least affected by data skewness among all methods compared with SVM. 

Transformer and XGB perform best at the higher and lower taxonomic levels, respectively. All classifiers perform worse at the lower taxonomic level, with mean score drops ranging from 9\% to 30\%. SVM is the most affected with a mean score drop of 26.1\%, opposite with XGB with a mean score drop of 9.74\%

\begin{table*}[]
\caption{Comparison of machine learning algorithms on data set 1}
\label{tab_comparison_ml}
\centering
\ra{1.3}
\resizebox{\textwidth}{!}{
\begin{tabular}{cccccccccc}
\toprule
  \textbf{Classifiers}                   & \multicolumn{4}{c}{\textbf{Higher Classification Level}}                                                    && \multicolumn{4}{c}{\textbf{Lower Classification Level}}                                                    \\
 \cline{2-5} 
\cline{7-10}
& \textbf{Mean Score (\%)} & \textbf{AUCPR (\%)} & \textbf{F1 (\%)} & \textbf{MCC (\%)} && \textbf{Mean Score (\%)} & \textbf{AUCPR (\%)} & \textbf{F1 (\%)} & \textbf{MCC (\%)} \\
\midrule
\textbf{CNN}         & 97.07 (0.74)              & 99.20 (0.31)      & 97.05 (0.71)      & 94.96 (1.22)       && 85.71 (2.01)              & 93.43 (1.31)      & 85.63 (1.87)      & 78.06 (2.86)       \\
\textbf{MLP}         & 93.00 (0.97)              & 96.63 (0.85)      & 93.49 (0.77)      & 88.88 (1.33)       && 78.10 (1.64)              & 87.42 (1.61)      & 78.95 (1.28)      & 67.94 (2.08)       \\
\textbf{RF}          & 97.41 (0.16)              & 99.31 (0.07)      & 97.38 (0.16)      & 95.52 (0.28)       && 87.57 (0.40)              & 94.53 (0.26)      & 87.43 (0.38)      & 80.75 (0.59)       \\
\textbf{RUSBoost}    & 59.49 (17.92)              & 72.80 (11.56)      & 60.74 (18.93)      & 44.92 (23.79)       && 47.96 (7.27)              & 55.97 (11.12)      & 51.88 (6.60)      & 36.02 (5.85)       \\
\textbf{SVM}         & 90.67 (3.78)              & 95.17 (2.33)      & 91.39 (3.51)      & 85.46 (5.55)       && 64.57 (15.01)              & 85.72 (3.63)      & 68.12 (11.21)      & 39.86 (30.68)       \\
\textbf{Transformer} & \textbf{97.86 (0.18) }             & \textbf{99.48 (0.06)}      & \textbf{97.82 (0.18)}      & \textbf{96.27 (0.32)}       && 87.33 (0.41)              & 94.55 (0.24)      & 87.11 (0.41)      & 80.34 (0.62)       \\
\textbf{XGB}         & 97.72 (0.18)              & 99.42 (0.07)      & 97.69 (0.18)      & 96.04 (0.31)       && \textbf{87.98 (0.52)}              & \textbf{94.83 (0.34)}      & \textbf{87.80 (0.49)}      & \textbf{81.31 (0.75)}      \\
\bottomrule
\end{tabular}
}
\end{table*}

\subsubsection{Overall Results\label{overall}}
\begin{figure*}[h]
\setlength{\abovecaptionskip}{0.cm}
\centering
\subfloat[Higher Classification Level]{%
  \includegraphics[width=0.5\textwidth]{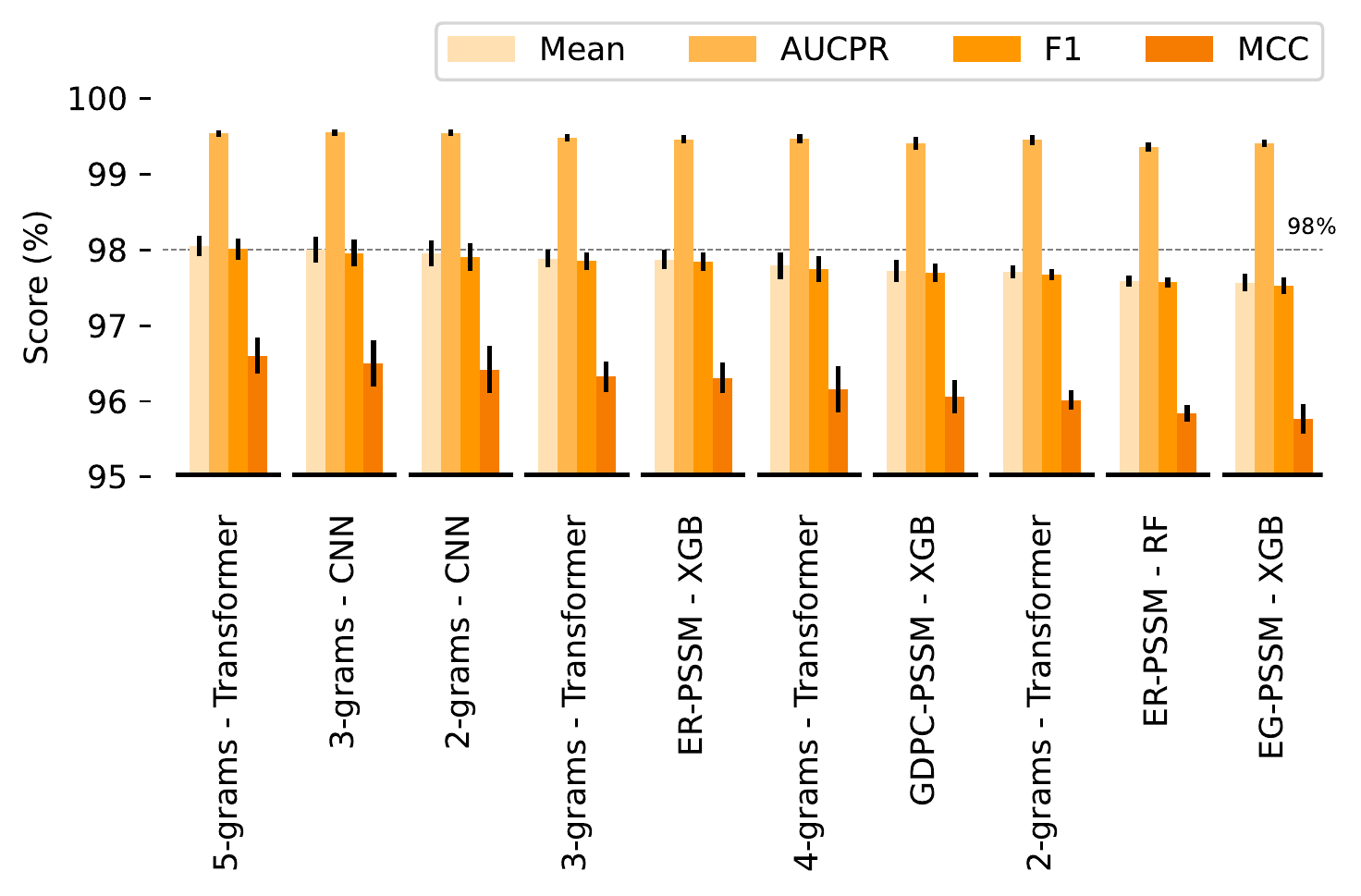}
}
\subfloat[Lower Classification Level]{%
  \includegraphics[width=0.5\textwidth]{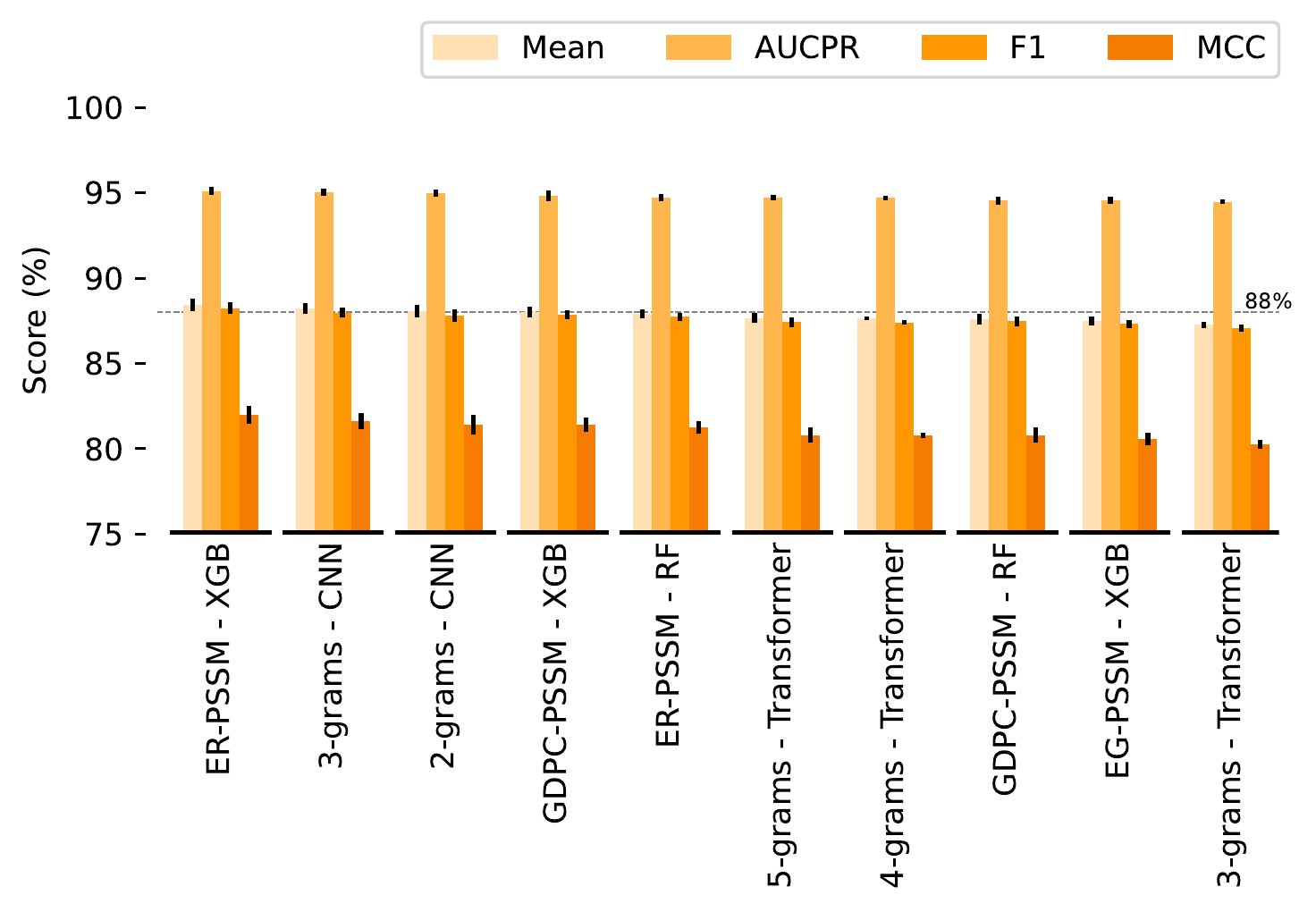}
}
\caption{Performance of different models at different classification levels on data set 1.}
\label{fig_hc_lc}
\end{figure*}

Fig.~\ref{fig_hc_lc} shows the top 10 models with the highest performance at different taxonomic levels. The results were ranked in descending order according to the mean score of each model. The name of the model is denoted as \textit{sequence representation - machine learning algorithm}. Most of the models reached at least a 96\% mean score when the data had fewer classes. ER-PSSM-XGB, 3-grams-CNN and 5-grams-Transformer work better at both taxonomic levels than other models. 

\subsection{Performance in Individual Hosts}
\begin{figure*}[h]
\setlength{\abovecaptionskip}{0.cm}
\centering
\includegraphics[width = \linewidth]{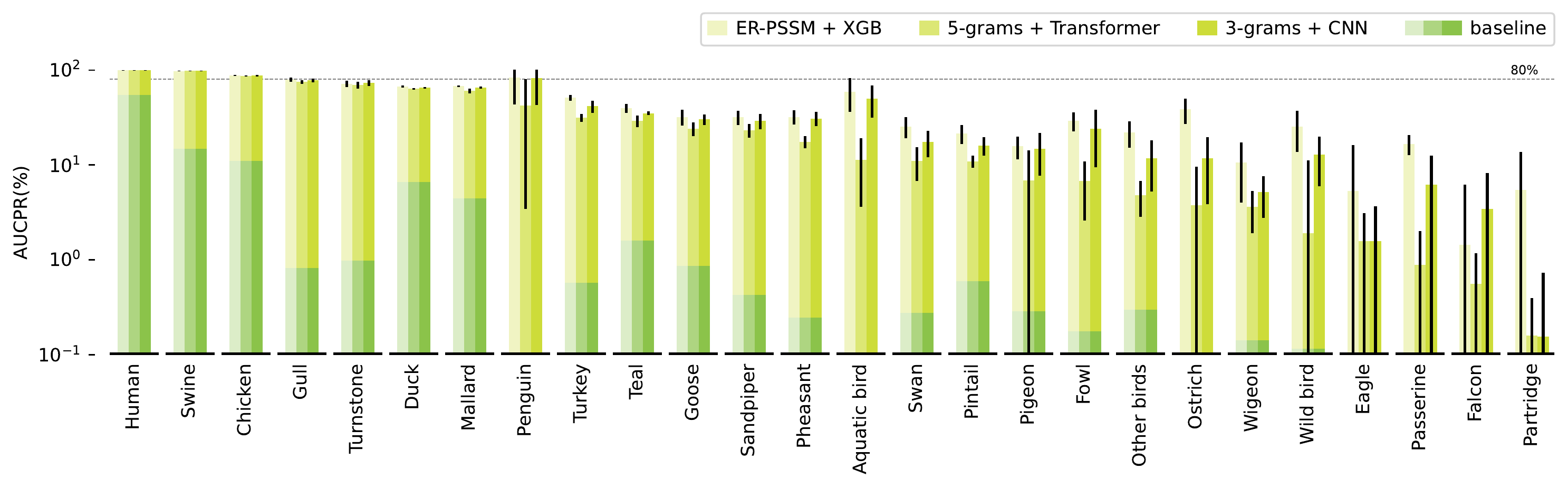}
\caption{Performance of different models in individual hosts at a lower taxonomic level on data set 1.}
\label{fig_individual_1}
\end{figure*}

Most machine learning algorithms focus on the majority class; therefore, it can be assumed they perform better on the majority class than on the minority class. In data set 1, 55\% of sequences belong to humans, while only 0.01\% of sequences belong to partridges. This degree of data imbalance brings great challenges to classifiers. ER-PSSM-XGB, 3-grams-CNN, and 5-grams-Transformer are the top three models that work better no matter how skewed the data. Their AUCPR score in individual hosts of data set 1 is shown in Fig.~\ref{fig_individual_1}. 

All three models scored AUCPR below 80\% in all hosts, with the exception of human, swine, and chicken which account for approximately 81\% of the sequences in data set 1.  However, the baseline of the AUCPR for each class is the proportion of each class in the data set. Therefore, human, swine, and chicken classes also have a relatively higher baseline than other classes. The AUCPR and corresponding baseline for each class are shown in lime and green. These three models achieved higher scores than the baseline, but the variability increased with fewer classes. 

\begin{figure*}[h]
\setlength{\abovecaptionskip}{0.cm}
\centering
\subfloat[Include Incomplete Sequences]{%
  \includegraphics[width=0.5\textwidth]{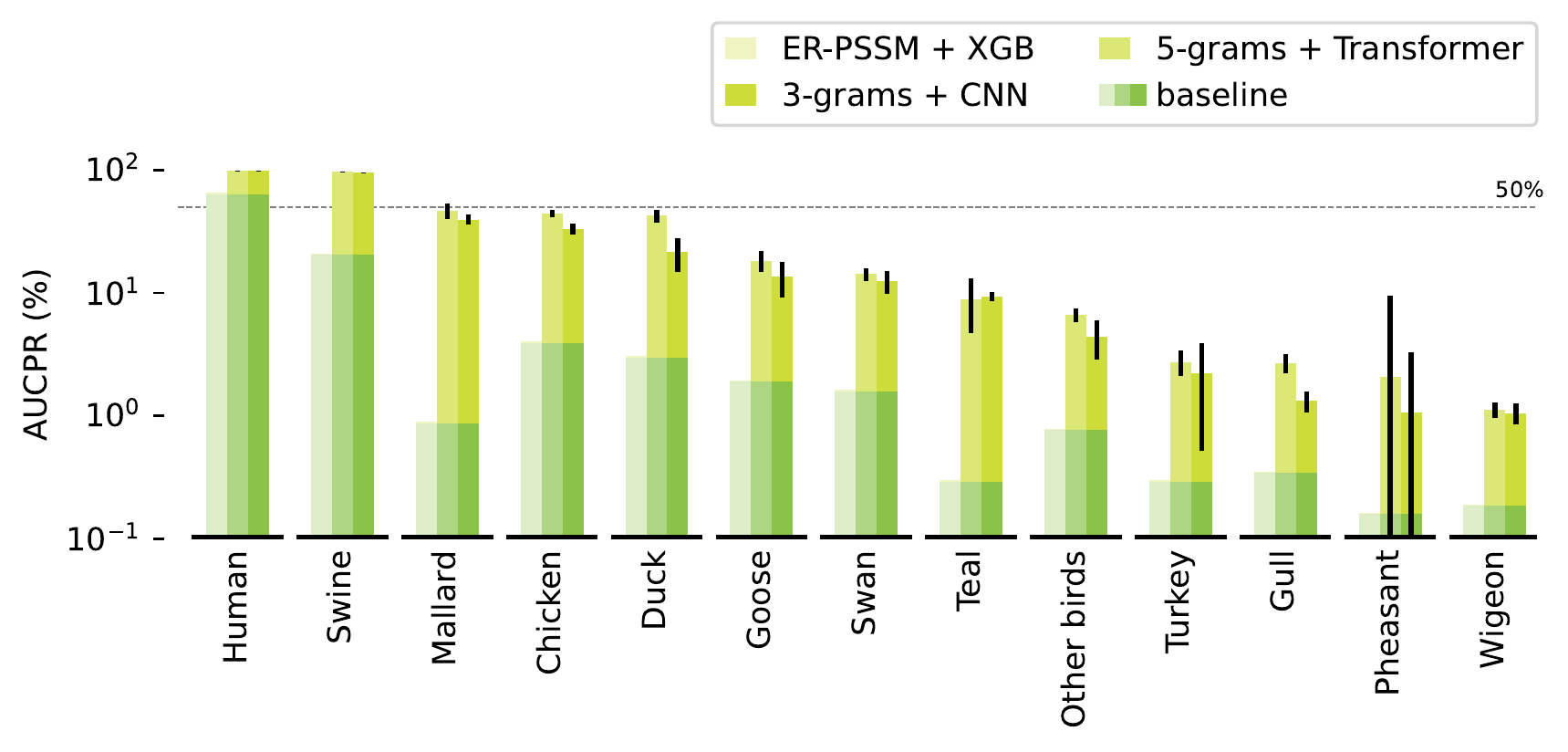}
}
\subfloat[Exclude Incomplete Sequences]{%
  \includegraphics[width=0.5\textwidth]{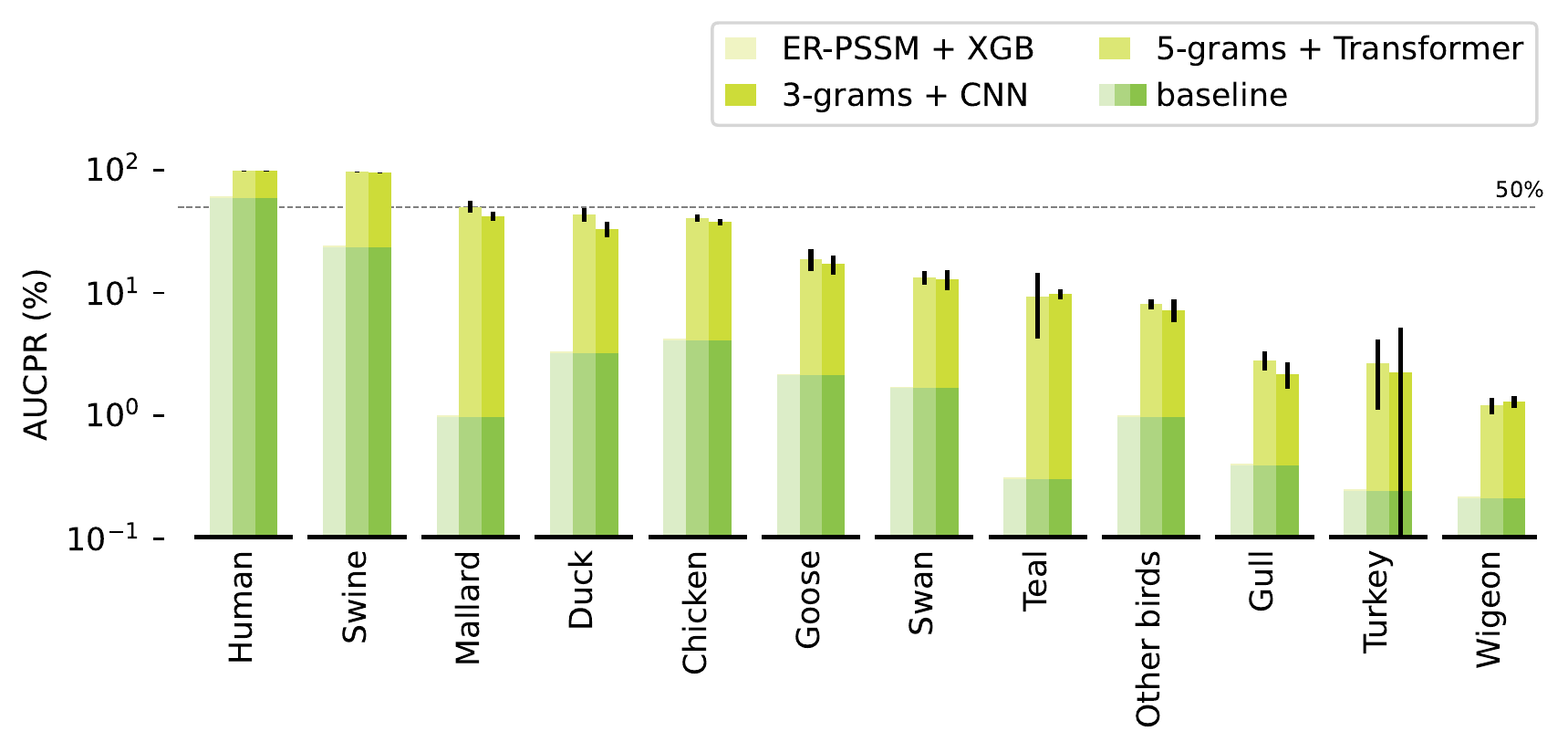}
}
\caption{Performance of different models in individual hosts at a lower taxonomic level on data set 2.}
\label{fig_individual_2}
\end{figure*}

Fig.~\ref{fig_individual_2} illustrates the performance of the three models in individual hosts at a lower taxonomic level for data set 2. The AUCPR of 5-grams-Transformer and 3-grams-CNN in each class still outperformed the baseline, even with the introduction of incomplete sequences, while ER-PSSM-XGB very slightly beat the baseline. 

\subsection{Effect of Incomplete Sequences}
\begin{figure*}[h]
\setlength{\abovecaptionskip}{0.cm}
\centering
\subfloat[Include Incomplete Sequences]{%
  \includegraphics[width=0.5\textwidth]{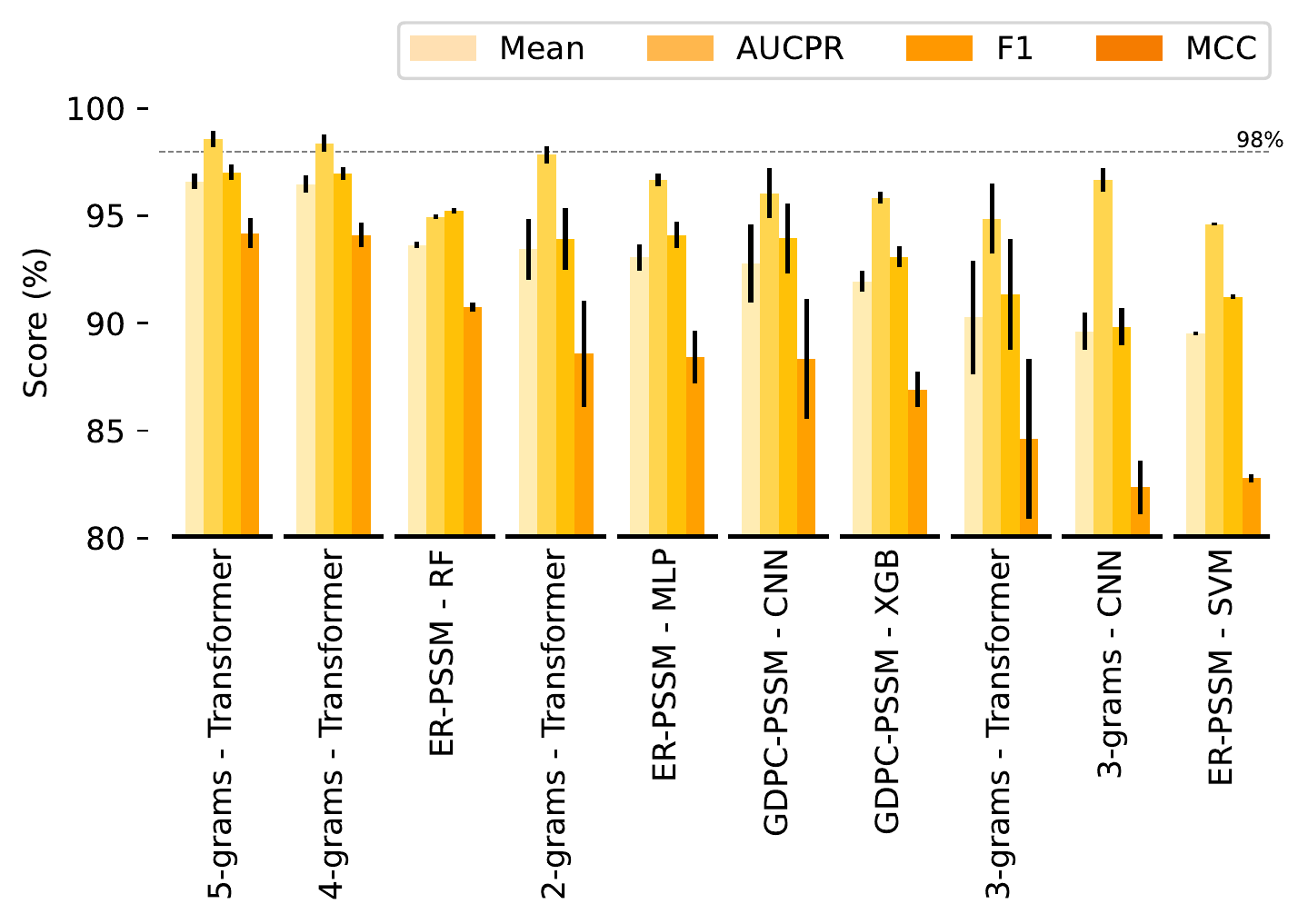}
}
\subfloat[Exclude Incomplete Sequences]{%
  \includegraphics[width=0.5\textwidth]{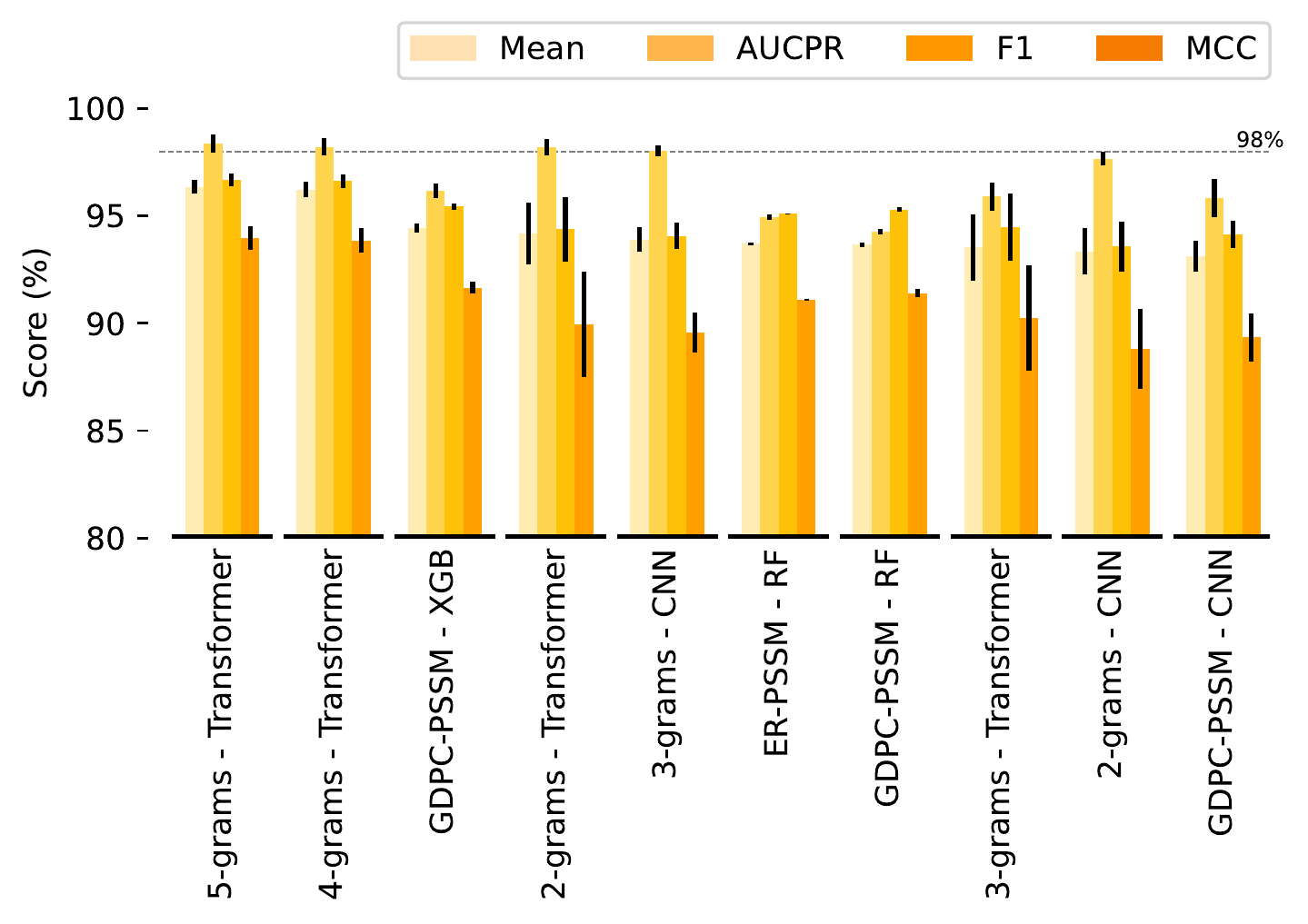}
}
\caption{Performance of different models at a higher taxonomic level on data set 2.}
\label{fig_test_hc}
\end{figure*}

\begin{figure*}[h]
\setlength{\abovecaptionskip}{0.cm}
\centering
\subfloat[Include Incomplete Sequences]{%
  \includegraphics[width=0.5\textwidth]{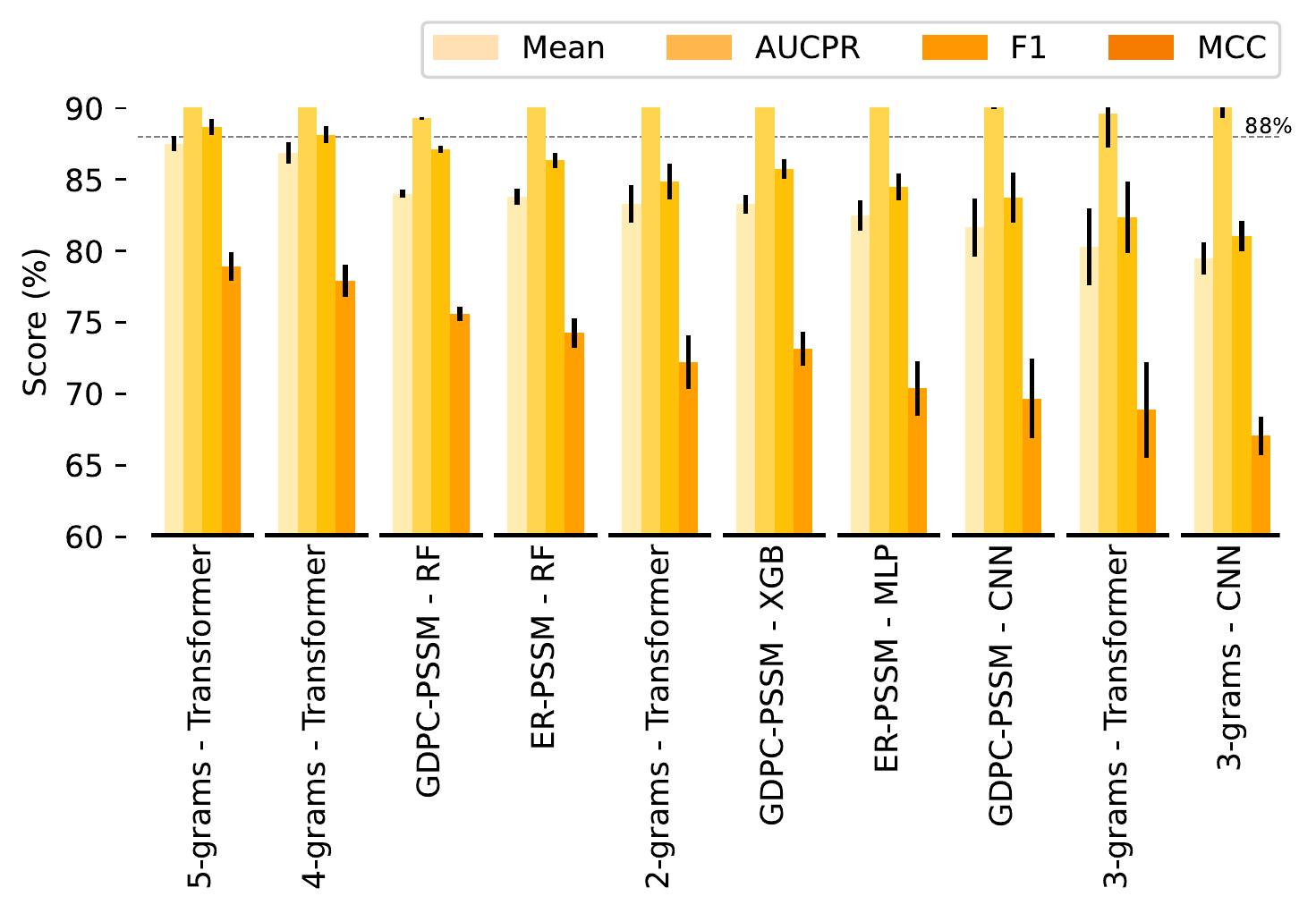}
}
\subfloat[Exclude Incomplete Sequences]{%
  \includegraphics[width=0.5\textwidth]{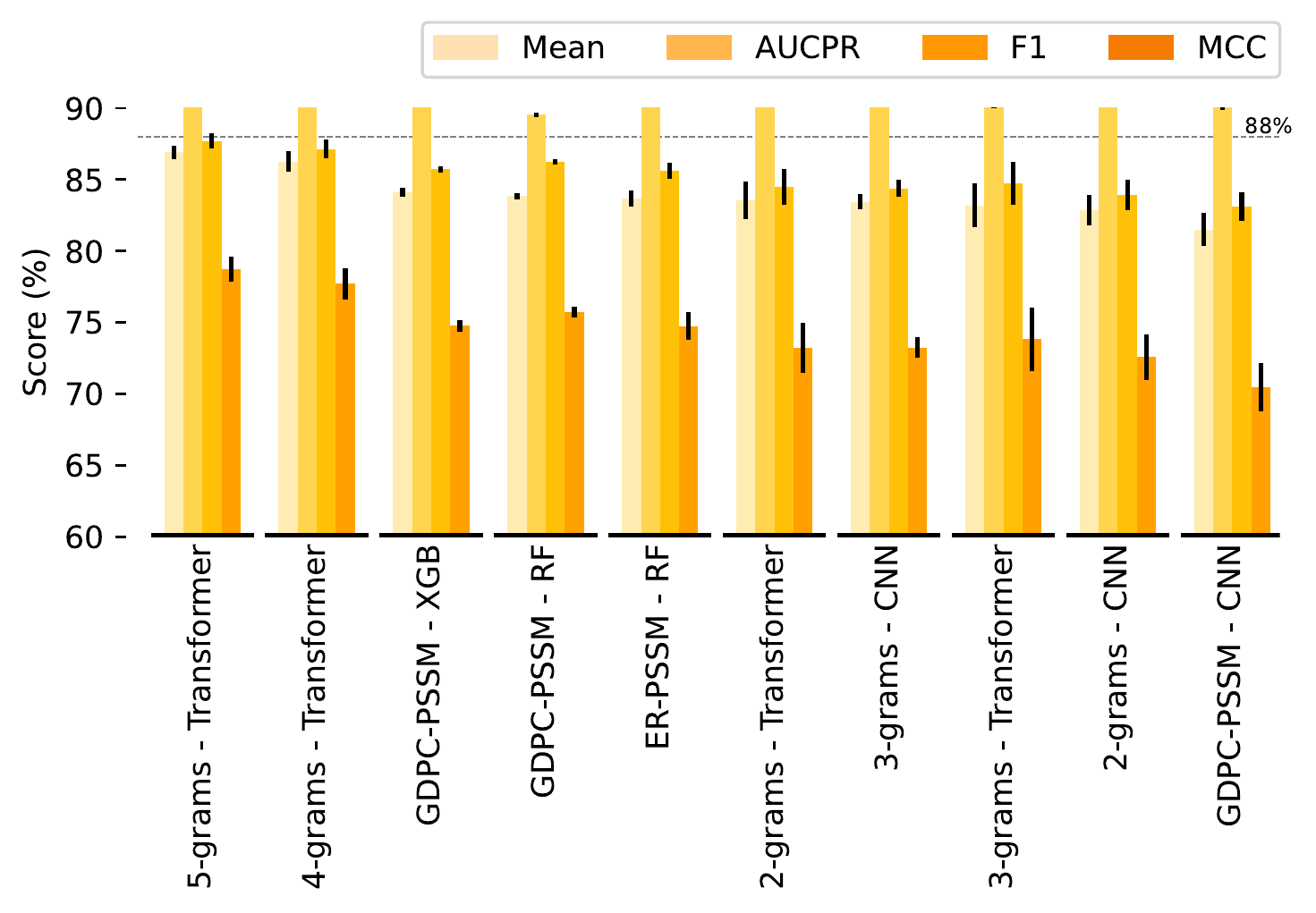}
}
\caption{Performance of different models at a lower taxonomic level on data set 2.}
\label{fig_test_lc}
\end{figure*}

Additionally, we investigated the effect of incomplete sequences on all models. Data set 1 contained numerous selected HA sequences that we used to produce a pre-trained model. Data set 2 contains 103 incomplete sequences, which we also used for evaluating the impact of incomplete sequences on models for data set 2, as shown in Fig.~\ref{fig_test_hc} and Fig.~\ref{fig_test_lc}.

The Transformer algorithm outperforms others at both taxonomic levels and is least affected by incomplete sequences. Overall, the performance of each model on data set 2 was reduced, but the impact of the incomplete sequence on the model was small.

\subsection{Ensemble Results}
We assembled all models and generated predictions for data set 1. The prediction results for 97.56\% of sequences matched the true label we assigned, while the prediction results for 2.44\% of sequences did not match. More specifically, the predictions results for 0.88\% of the sequences from all models did not match the true labels. To determine which sequences were not predicted “correctly” by all models, we performed a basic statistical analysis. 

\begin{figure}[h]
\setlength{\abovecaptionskip}{0.cm}
\centering
\includegraphics[width = 0.5\linewidth]{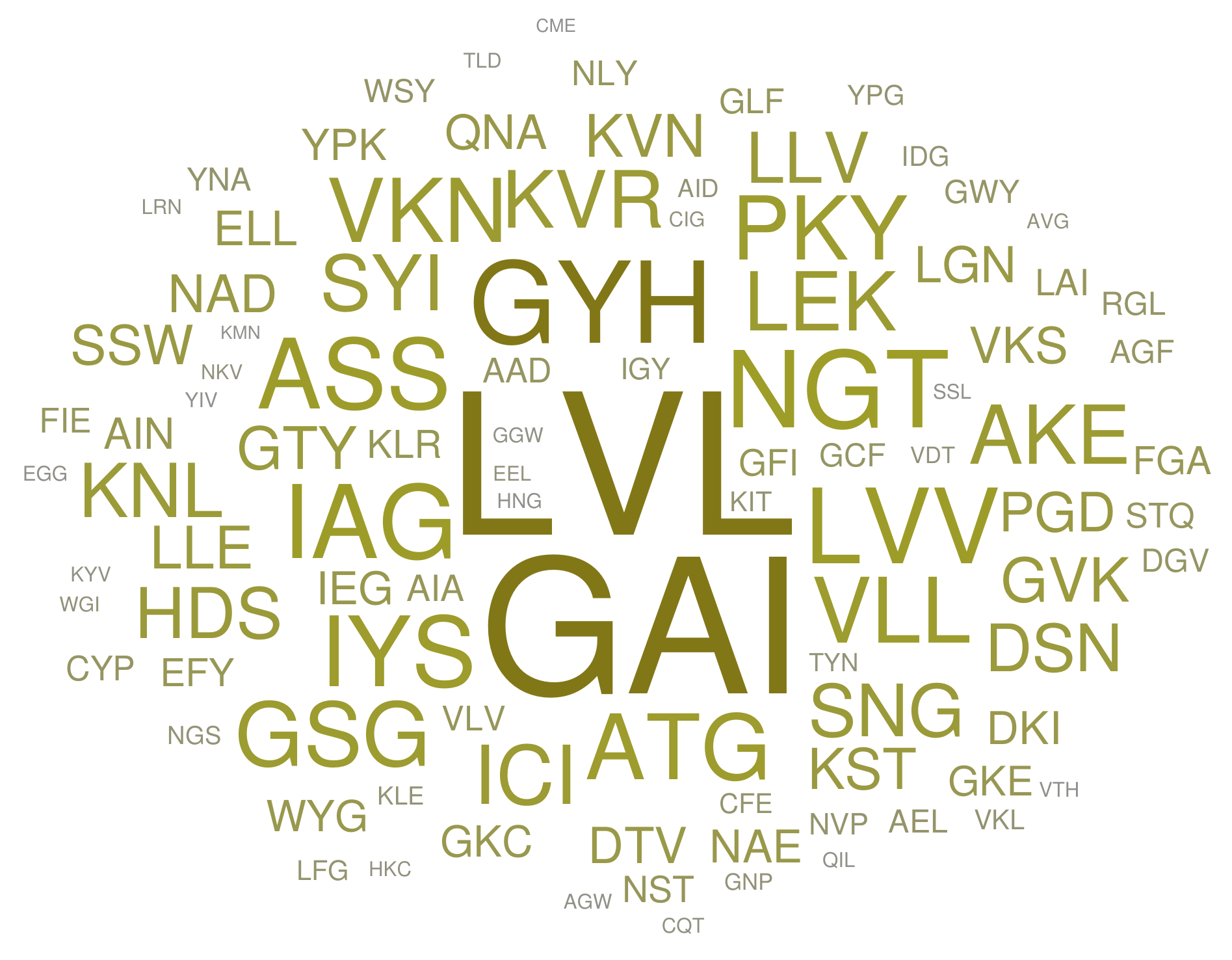}
\caption{Word clouds of sequences that cannot be "correctly" predicted by all models, generated by MATLAB\textsuperscript{®}}
\label{fig_ensemble}
\end{figure}

Approximately 30\% of the collected sequences were collected between 2009 and 2011, when the 2009 Influenza pandemic occurred; 57.52\% and 40.76\% of the sequences belong to swine and human; and 50.66\%, 23.42\%, 8.57\%, and 4.00\% of the sequences belong to H1N1, H5N1, H9N2, and H5N6. We also plotted the word clouds for these sequences, as shown in  Fig.~\ref{fig_ensemble}. The most frequent tokens revealed were LVL and GAI, which were also the most frequent tokens in swine sequences (Fig.~\ref{fig_wordcloud}).

Sequences that failed to fit the model may have been collected during outbreaks (e.g. pH1N1-like) or have strong cross-species capabilities. For example, \emph{A/Beijing/1/2017} and \emph{A/India/TCM2581/2019} are avian viruses isolated from human \citep{pan2018human}, \citep{potdar2019laboratory}; therefore, the label of these sequences is human while our prediction is avian. Similarly, \emph{A/swine/Jangsu/48/2010} is a pH1N1-like swine virus used to prove retro-infection from swine to human in China \citep{zhao2012isolation}. This sequence is labelled swine while our prediction is human. 

\section{Discussion and Conclusion\label{discussion}}
In this work, we applied popular supervised machine learning algorithms to predict Influenza A virus hosts given hemagglutinin sequences. Popular classic machine learning and deep learning algorithms were evaluated.  One difference between classic machine learning and deep learning is how the data is interpreted. Machine learning cannot interpret raw data directly. Therefore, both forms require numerical input data. Typical machine learning typically requires a feature extraction step, while deep learning analyses the features of the input data through hidden layers.

We implemented a positional-specific scoring matrix (PSSM) to extract evolutionary features of sequences, and then fed them into classic machine learning algorithms (SVM, RF, RUSBoost, and XGB). We proposed three methods to unify the dimension of each PSSM (i.e., ER, GDPC, and EG). For deep learning models (i.e., Transformer, CNN, and MLP), an embedding layer can be added to learn features automatically. Without an embedding layer, a feature extraction step should be applied before feeding the input data into models. For large data sets, the computation of PSSMs is time-consuming and laborious, taking days or even months. To reduce computational time, we can reduce the size of the data set or perform a BLAST search using a partial protein database, but at the expense of reduced accuracy. The other disadvantage of using PSSMs is that they do not have a unified dimension, so unifying strategies are required to process PSSMs in a more efficient manner. Therefore, end-to-end deep learning approaches address these limitations without the need for cumbersome feature engineering. In this study, we demonstrate that word embedding can be a useful alternative to processing sequences. 

To evaluate the models on different taxonomic levels and different data sets, nested cross-validation and a variety of evaluation metrics were used. Sequences in data set 1 were chosen selectively, including only unique complete sequences. The sequences in data set 2 were collected in the most recent year, with only redundant sequences discarded. The two data sets are mutually exclusive and the model is blind to data set 2. Based on our results, ER-PSSM-XGB has the best performance in data set 1 at a lower classification level, whereas 5-grams-Transformer performs optimally at a higher classification level. 

Data set 2 contains a small proportion of incomplete sequences, which can be assumed as noisy data or data with incomplete information. In this circumstance, the 5-gram-Transformer still performs optimally. The Transformer is more powerful than classic RNNs and CNNs because it incorporates the multi-head attention mechanism that focuses on the most important part of the sequence and introduces the information of sequence order by positional encoding, which is more powerful than classic RNNs.

We exclusively utilised supervised machine learning but supervised learning relies on ground truth. We extracted class labels for sequences from metadata, which indicated the isolated host of sequences. As a result, some labels used in this study do not perfectly represent the ground truth. In contrast to supervised learning, semi-supervised or unsupervised learning requires partial or even no ground truth during training and can be combined in future studies.

\section*{Acknowledgement}
This work is supported by the University of Liverpool.


\nocite{*}
\bibliographystyle{cas-model2-names}

\bibliography{cas-refs}

\end{document}